\documentclass{article}

\usepackage{microtype}
\usepackage{graphicx}
\usepackage{subfigure}
\usepackage{booktabs} % for professional tables
\usepackage{colortbl}
\usepackage{hyperref}

\usepackage{multirow}
\newcommand{\h}[1]{\mathbf{h}_{#1}} % the representation
\newcommand{\z}[1]{\mathbf{z}_{#1}} % the projection
\newcommand{\X}[1]{\mathbf{X}_{#1}} 
\newcommand{\MP}[1]{\mathbf{P}_{#1}} 
\newcommand{\B}[1]{\mathbf{B}_{#1}} 
\newcommand{\MS}[1]{\mathbf{S}_{#1}}
\newcommand{\MA}[1]{\mathbf{A}_{#1}} 
\newcommand{\MH}[1]{\mathbf{H}_{#1}} % the projection
\newcommand{\Z}[1]{\mathbf{Z}_{#1}} % the projection

\newcommand{\Term}[1]{\textit{#1}}
\usepackage[accepted]{icml2025}
\usepackage{marvosym}

\usepackage{utfsym}
\usepackage{amsmath}
\usepackage{amssymb}
\usepackage{mathtools}
\usepackage{amsthm}
\usepackage{arydshln}
\usepackage[capitalize,noabbrev]{cleveref}
\usepackage{graphicx}
\theoremstyle{plain}
\newtheorem{theorem}{Theorem}[section]

\theoremstyle{definition}
\newtheorem{definition}[theorem]{Definition}

\theoremstyle{remark}

\usepackage[textsize=tiny]{todonotes}

\icmltitlerunning{Clustering Properties of Self-Supervised Learning}

\begin{document}

\twocolumn[
\icmltitle{Clustering Properties of Self-Supervised Learning}



\icmlsetsymbol{equal}{*}

\begin{icmlauthorlist}
\icmlauthor{Xi Weng}{buaa,nus} \quad \quad
\icmlauthor{Jianing An}{buaa} \quad \quad
\icmlauthor{Xudong Ma}{buaa} \quad \quad
\icmlauthor{Binhang Qi}{nus} \\ \quad \quad
\icmlauthor{Jie Luo}{buaa} \quad \quad 
\icmlauthor{Xi Yang}{baai} \quad \quad  
\icmlauthor{Jin Song Dong}{nus} \quad \quad
\icmlauthor{Lei Huang$^{~\textrm{\Letter}}$}{buaa,hz} \quad \quad
\end{icmlauthorlist}

\icmlaffiliation{buaa}{SKLCCSE, School of Artificial Intelligence, Beihang University}
\icmlaffiliation{nus}{School of Computing, National University of Singapore}
\icmlaffiliation{baai}{Beijing Academy of Artificial Intelligence}
\icmlaffiliation{hz}{Hangzhou International Innovation Institute, Beihang University}
\icmlcorrespondingauthor{Lei Huang}{huangleiAI@buaa.edu.cn}

\icmlkeywords{Machine Learning, ICML}

\vskip 0.2in

]


\printAffiliationsAndNotice{}  % leave blank if no need to mention equal contribution

\begin{abstract}
Self-supervised learning (SSL) methods via joint embedding architectures have proven remarkably effective at capturing semantically rich representations with strong clustering properties, magically in the absence of label supervision. Despite this, few of them have explored leveraging these untapped properties to improve themselves. In this paper, we provide an evidence through various metrics that the encoder's output \Term{encoding} exhibits superior and more stable clustering properties compared to other components. Building on this insight, we propose a novel positive-feedback SSL method, termed \textbf{Re}presentation \textbf{S}elf-\textbf{A}ssignment (ReSA), which leverages the model's clustering properties to promote learning in a self-guided manner. Extensive experiments on standard SSL benchmarks reveal that models pretrained with ReSA outperform other state-of-the-art SSL methods by a significant margin. Finally, we analyze how ReSA facilitates better clustering properties, demonstrating that it effectively enhances clustering performance at both fine-grained and coarse-grained levels, shaping representations that are inherently more structured and semantically meaningful.
\end{abstract}

\vspace{-0.3in}
\section{Introduction}
\label{introduction}

    \begin{figure}[ht]
        \vspace{-0.05in}
        \begin{center}   \centerline{\includegraphics[width=5.7cm]{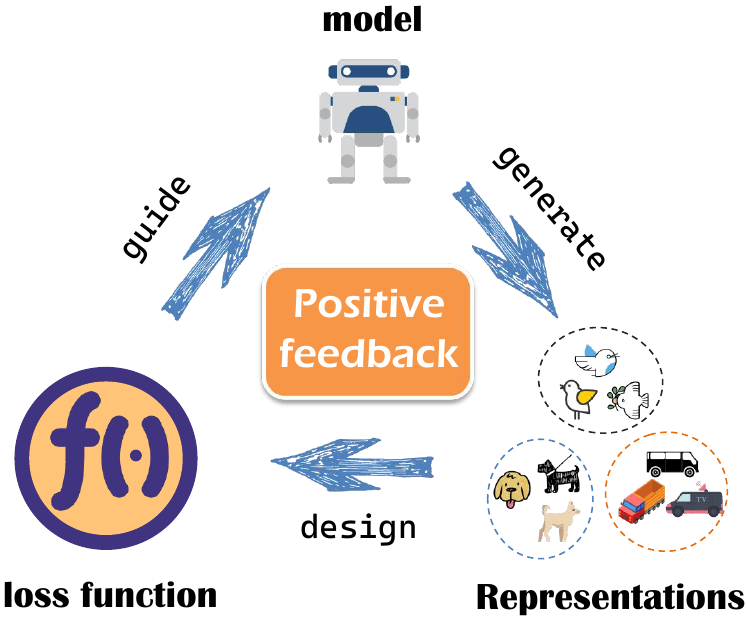}}
        \vskip -0.1in
        \caption{The positive-feedback SSL framework. It involves the model generating representations that possess semantically clustering information. This clustering information is leveraged to design self-supervised loss function, which is then employed to more effectively guide the model's learning process.}
    \label{fig:pf}
    \vspace{-0.4in}
    \end{center}
    \end{figure}
\vspace{-0.05in}
Self-supervised learning (SSL) has emerged as a transformative paradigm in universal representation learning~\cite{2018_arxiv_Oord,2019_NIPS_Bachman,2020_CVPR_He,2020_ICML_Chen,bao2021beit,2023_arxiv_Oquab,2023_cvpr_Assran}, consistently surpassing supervised learning in downstream  performance. Joint embedding architectures (JEA), in particular, aim to learn invariance of the same data under different transformations and noise~\cite{2019_NIPS_Bachman,2020_CVPR_He,2020_ICML_Chen}, with demonstrated exceptional effectiveness in visual representation learning. Although such a pretext task may intuitively seem unrelated to capturing semantic relationships, extensive studies~\cite{2018_ECCV_Caron,2021_ICCV_Caron,2022_ECCV_Assran} have demonstrated the strong correlation between its learned representations and semantic information.

\vspace{-0.05in}
\citet{2023_NIPS_Ben-Shaul} take a further step to characterize the semantic structures learned by JEA into hierarchic clustering properties, indicating SSL-trained representations exhibit a centroid-like geometric structure and induce three levels of semantic clustering: augmentation sample level, semantic classes, and superclass level. This intriguing finding reveals that SSL methods based on JEA can facilitate strong clustering capabilities during training, but also raises the question of \textit{whether these properties hold untapped potential that can be further leveraged to improve SSL itself}.
\vspace{-0.27in}
\paragraph{Contributions.} In this paper, we aim to investigate the design of SSL methods by leveraging the inherent clustering properties of representations, enabling a closed-loop positive-feedback SSL framework, as illustrated in Figure~\ref{fig:pf}. To achieve this goal, we propose three key questions and our main contributions are summarized as follows:
\begin{itemize}
    \vspace{-0.155in}
    \item \textbf{Where to extract clustering properties from?} We demonstrate through various metrics that the encoder's output, referred to as \Term{encoding}, exhibits superior and more stable clustering properties compared to other components, such as \Term{embedding} and the \Term{hidden layer outputs} within the projector. 
    
    \vspace{-0.075in}
    \item \textbf{How to leverage clustering properties?} We propose a novel SSL method, termed \textbf{Re}presentation \textbf{S}elf-\textbf{A}ssignment (ReSA), which employs an online self-clustering mechanism to leverage the model's inherent clustering properties, thereby facilitating positive-feedback learning. Standard experiments demonstrate that ReSA surpasses existing state-of-the-art methods in both performance and training efficiency.
    
    \vspace{-0.05in}
    \item \textbf{Whether it facilitates better clustering properties?} We examine whether and how ReSA facilitates better clustering properties, demonstrating that it excels at both fine-grained and coarse-grained learning, shaping representations that are inherently more structured and semantically meaningful.  
\end{itemize}

\vspace{-0.2in}
\section{Background, Related Work, $\&$ Notation}

\subsection{Self-Supervised Learning}
    \begin{figure}[ht]
    \begin{center}
    \vspace{-0.1in}
    \centerline{\includegraphics[width=6.5cm]{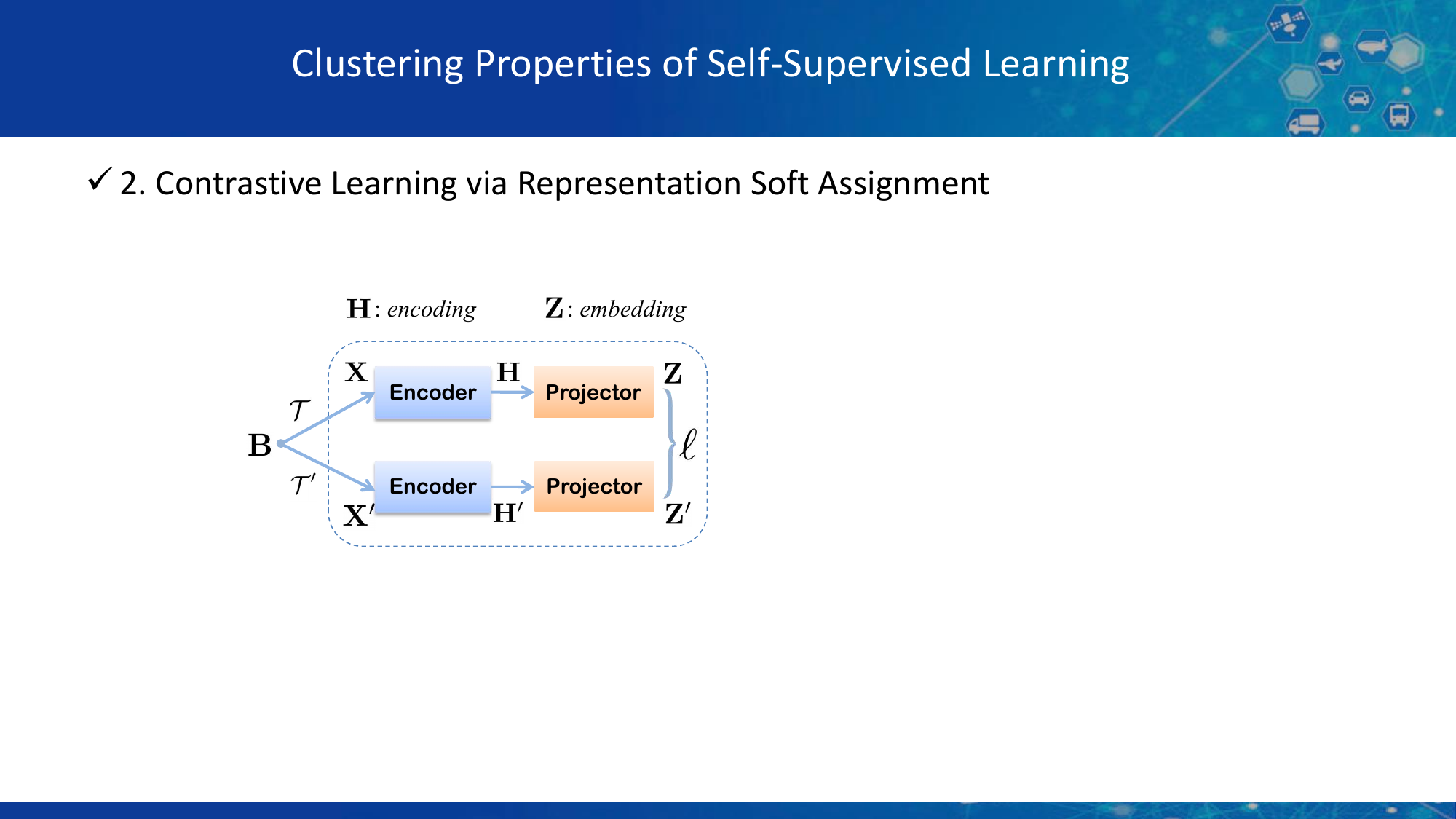}}
    \vspace{-0.15in}
    \caption{The basic notations for joint embedding architectures (JEA) in SSL.}
    \label{fig:JEA}
    \end{center}
    \vspace{-0.4in}
    \end{figure}
    
\paragraph{Joint embedding architectures (JEA).}
\label{para:jea}
	Let $\B{}$ denote a mini-batch input sampled uniformly from a set of images $\mathbb{D}$, and $\mathbb{T}$  denote the set of data transformations available for augmentation.  We consider a pair of neural networks $F_{\theta}$ and $F'_{\theta'}$, parameterized by $\theta$  and $\theta'$ respectively. They take as input two randomly augmented views, $\X{} = \mathcal{T}(\B{}) $ and $\X{}'=\mathcal{T}'(\B{})$, where $\mathcal{T}, \mathcal{T}'\in \mathbb{T}$; and they output the \Term{embeddings} $\Z{} =F_{\theta}(\X{})$ and $\Z{}'=F'_{\theta'}(\X{}')$. The networks are trained with an objective function that minimizes the distances between \Term{embeddings} obtained from different (two) views of the same images:
	\begin{eqnarray}
		\label{eqn:objective}
		\mathcal{L}(\B{},\theta)= \mathbb{E}_{\B{} \sim \mathbb{D},~\mathcal{T}, \mathcal{T}' \sim \mathbb{T}}~~ \ell \big(F_{\theta} (\mathcal{T}(\B{})), F'_{\theta'}(\mathcal{T}'(\B{}))\big).
	\end{eqnarray}
where $\ell(\cdot, \cdot)$ is a loss function, which aims to learn invariance of  data transformations.

In particular, the JEA usually consist of a shared encoder $E_{\theta_e}(\cdot)$ and projector $G_{\theta_g}(\cdot)$, commonly referred to as a Siamese Network~\cite{2021_CVPR_Chen}. As shown in Figure~\ref{fig:JEA}, their outputs $\MH{}=E_{\theta_e}(\X{}) \in  \mathbb{R}^{d_e \times m }$ and $\Z{}=G_{\theta_g}(\MH{}) \in  \mathbb{R}^{d_g \times m }$ are referred to as \Term{encoding} and \Term{embedding}, respectively (where $m$ is the mini-batch size, $d_e$ and $d_g$ are their corresponding feature dimensions). Under this notation, we have $F_{\theta}(\cdot) = G_{\theta_g}(E_{\theta_e}(\cdot))$ with learnable parameters $\theta=\{\theta_e, \theta_g\}$. It is worth noting that extensive work~\cite{2020_ICML_Chen, 2022_projector_Gupta} has demonstrated that using the \Term{encoding} as the representation for downstream tasks achieves much better performance than using \Term{embedding}, so that the projector is only used during the pre-training process and discarded in inference.

\begin{figure}[t]
%\vskip 0.2in
\begin{center}
\centerline{\includegraphics[width=7.5cm]{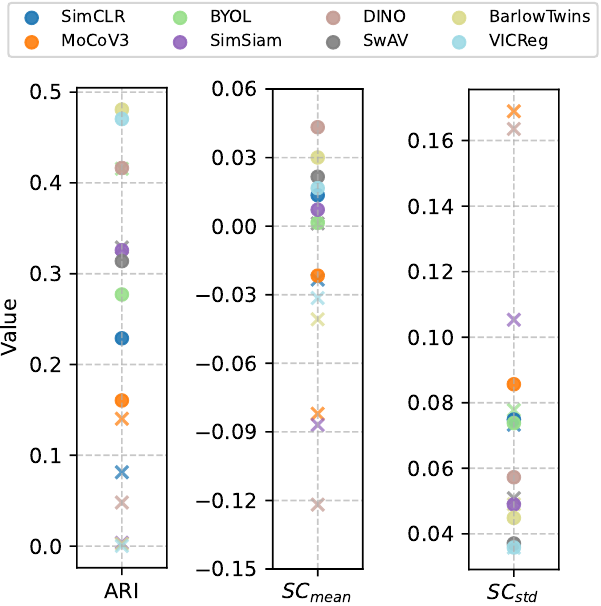}}
\vspace{-0.1in}
\caption{Comparison of clustering metrics in \Term{encoding} $\MH{}$ and \Term{embedding} $\Z{}$ across various self-supervised pretrained models. All methods utilize a ResNet-18 encoder pretrained on CIFAR-10 for 1000 epochs. Circular markers represent metrics computed using \Term{encodings}, while cross markers correspond to metrics derived from \Term{embeddings}. All metrics are computed on the entire training set, and similar trends can be observed in the validation set.}
\label{fig:rep_cifar10}
\end{center}
\vspace{-0.4in}
\end{figure}

\begin{figure*}[t]
%\vskip 0.2in
\begin{center}
\centerline{\includegraphics[width=16cm]{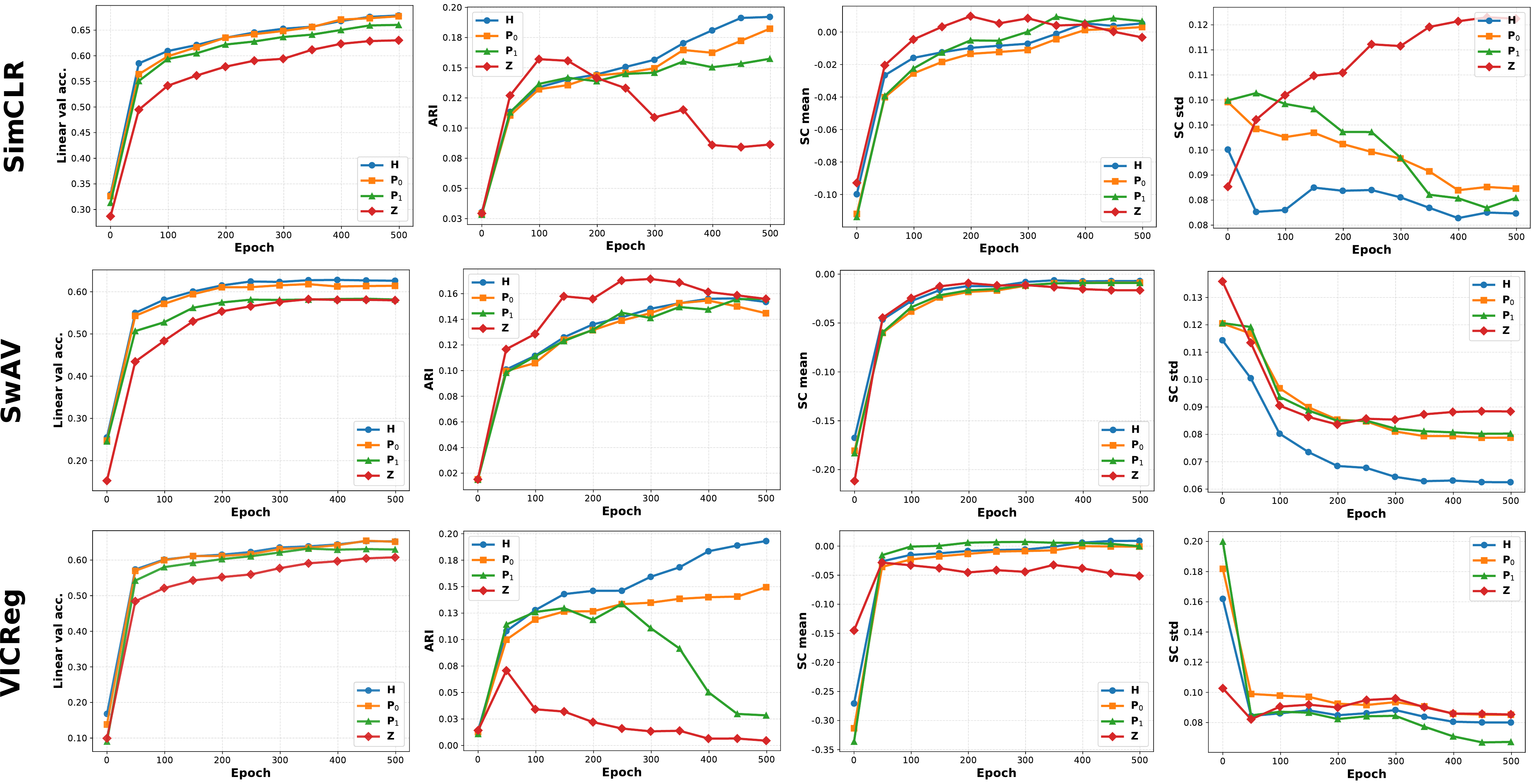}}
\vspace{-0.15in}
\caption{Comparison of linear evaluation accuracy and clustering metrics of \Term{encoding} $\MH{}$, \Term{embedding} $\Z{}$, and the \Term{hidden layer outputs within the projector} $\MP{}$ during the training process. The experiments are conducted using SimCLR, VICReg, and SwAV, employing a ResNet-18 encoder pretrained on CIFAR-100 for 500 epochs. The projector is a standard three-layer MLP with BN and ReLU activations, containing two hidden linear layers, so their outputs are denoted as $\MP{0}$ and $\MP{1}$.}
\vspace{-0.35in}
\label{fig:rep_cifar100}
\end{center}
\end{figure*}
\vspace{-0.18in}
\paragraph{Self-supervised paradigms.} The main challenge with JEA is representation collapse, where both branches produce identical and constant outputs regardless of the inputs. Numerous paradigms have been proposed to avoid collapse, including contrastive learning methods~\cite{2020_ICML_Chen,2020_CVPR_He,2020_arxiv_Chen,2021_mocov3_Chen,2022_NIPS_Liu} that attract different views from the same image (positive pairs) while pushing apart different images (negative pairs), and non-contrastive approaches~\cite{2020_NIPS_Grill,2020_NIPS_Caron,2021_ICCV_Caron,2024_ICLR_Weng} which directly align positive targets without incorporating negative pairs. Although ~\citet{2023_NIPS_Ben-Shaul} had demonstrated that the \Term{encodings} learned through SSL are highly correlated with semantic classes and exhibit strong clustering capabilities, few methods have leveraged this clustering ability to facilitate positive-feedback learning. A closely related work~\cite{2023_arxiv_Ma} exploits the \Term{encoding}'s augmentation robustness to re-weight the positive alignment in the SSL objective functions; however, it still overlooks the rich clustering information inherent in the \Term{encodings}.

%e.g. SimCLR~\cite{2020_ICML_Chen}, MoCos~\cite{2020_CVPR_He,2020_arxiv_Chen,2021_ICCV_Chen}, NNCLR~\cite{2021_ICCV_Dwibedi}, MEC~\cite{2022_NIPS_Liu} that attract different views from the same image (positive pairs) while pushing apart different images (negative pairs), and non-contrastive approaches, e.g. asymmetric (BYOL~\cite{2020_NIPS_Grill}, SimSiam~\cite{2021_CVPR_Chen}, DINO~\cite{2021_ICCV_Caron}), clustering (DeepCluster~\cite{2018_ECCV_Caron}, SwAV~\cite{2020_NIPS_Caron}), whitening methods (W-MSE~\cite{2021_ICML_Ermolov}, Barlow Twins~\cite{2021_NIPS_Zbontar}, VICReg~\cite{2022_ICLR_Adrien}, INTL~\cite{2024_ICLR_Weng}) that directly align positive targets without incorporating negative pairs. 

\subsection{The Information Distinction in JEA} 
\vspace{-0.05in}

The projector has become an indispensable component of JEA-based SSL. However, the theoretical dynamics of its optimization and the reasons behind its success remain an open question within the community. Some works have attempted to explain these principles. For example, \citet{2022_ICLR_Jing} empirically discovered that applying SSL loss either to \Term{encoding} or \Term{embedding} led to a significant decrease in the rank of the corresponding features. They argued that this rank reduction indicates a loss of diverse information, which, in turn, reduces generalization capability. This explanation aligns with the hypothesis in SimCLR~\cite{2020_ICML_Chen}, where the additional projector acts as a buffer to prevent information degradation of the \Term{encoding} caused by the invariance constraint. Additionally, \citet{2022_projector_Gupta}'s null space analysis for the projector posited that the projector might implicitly learn to select a subspace of the \Term{encoding}, which is then mapped into the \Term{embedding}. In this way, only a subspace of the \Term{encoding} is encouraged to be style-invariant, while the other subspace can retain more useful information.

Therefore, the SSL constraint can cause the \Term{embedding} to lose information, which may include not only clustering-irrelevant features such as background information, but also class-relevant information, making it difficult to determine which—\Term{encoding} or \Term{embedding}—exhibits better clustering performance in this context. In such cases, this paper first analyzes the differences between the two in terms of clustering properties empirically.

\section{Exploring Clustering Properties of SSL}
\label{sec:properties}
\vspace{-0.05in}
\citet{2023_NIPS_Ben-Shaul} find that within the encoder of JEA, the clustering ability of features improves progressively as intermediate layers get deeper. However, it remains unclear whether the projector exhibits a similar trend. To quantitatively evaluate the clustering performance of these components, we employ widely recognized metrics such as the Silhouette Coefficient (SC)~\cite{1987_JCAM_Rousseeuw} and Adjusted Rand Index (ARI)~\cite{1985_JC_Hubert}. In particular, a larger mean value of SC ($\text{SC}_{\text{mean}}$) indicates stronger \textbf{local} clustering ability in the representation, and a smaller standard deviation ($\text{SC}_{\text{std}}$) reflects better stability in local clustering~\footnote{The `clustering ability' refers to how well the vectors can represent the underlying structure of the data, while `stability' signifies that the clustering results are more consistent across the data points, meaning fewer outliers and more stable cluster assignments.}. Meanwhile, higher ARI values correspond to enhanced \textbf{global} clustering properties. The detailed introduction of these metrics can be found in the Appendix~\ref{appendix:metrics}.

Using these metrics, we first evaluate the clustering abilities of \Term{encoding} $\MH{}$ and \Term{embedding} $\Z{}$ across various self-supervised pretrained models in CIFAR-10~\cite{2009_TR_Alex} dataset, which contains only 10 classes and is commonly used for cluster analysis~\cite{2023_NIPS_Ben-Shaul}. It is evident in Figure~\ref{fig:rep_cifar10} that, across most SSL models, \Term{encodings} achieve visibly higher ARI and $\text{SC}_{\text{mean}}$, as well as lower $\text{SC}_{\text{std}}$ values, compared to \Term{embeddings}. These observations reflect common grounds of SSL models: \Term{Encodings} not only possess richer semantic information but also demonstrate high-quality clustering properties. These include excellent local clustering ability ($\text{SC}_{\text{mean}}$) and stability ($\text{SC}_{\text{std}}$), global clustering capability and similarity measure effectiveness (ARI). An exception to this pattern is observed with SwAV and BYOL, whose \Term{embeddings} also perform well across these metrics. We speculate that this may be due to the design of their loss functions, which enables the \Term{embeddings} to learn effective clustering properties. For instance, SwAV learns by predicting cluster assignments directly on the \Term{embeddings}.

\vspace{-0.05in}
To gain deeper insights into the evolution of clustering properties of each component during training, we conduct experiments on the CIFAR-100~\cite{2009_TR_Alex} dataset, which features a more complex set of categories, using three methods: SimCLR, VICReg, and SwAV. The results are shown in Figure~\ref{fig:rep_cifar100}. Overall, the \Term{encodings} demonstrate a consistent improvement in all clustering metrics throughout training. In contrast, the ones of \Term{embeddings} degrade significantly during the later stages of training. Moreover, the clustering metrics of the \Term{hidden layer outputs within the projector} show notable differences and weaknesses compared to those of the \Term{encodings}, despite $\MP{0}$ being separated by only a single linear layer and achieving nearly the same linear evaluation accuracy as the \Term{encodings}.

\vspace{-0.05in}
In summary, the experiments above demonstrate that \Term{encodings} exhibit superior and more stable clustering properties compared to \Term{embeddings} and the \Term{hidden layer outputs within the projector} across various SSL models. This finding highlights the potential of leveraging \Term{encoding} as the optimal representation for clustering, providing a foundation for designing positive-feedback SSL systems that capitalize on these robust clustering properties.

\begin{figure}[t]
    %\vskip 0.2in
    \begin{center}
    \centerline{\includegraphics[width=8.2cm]{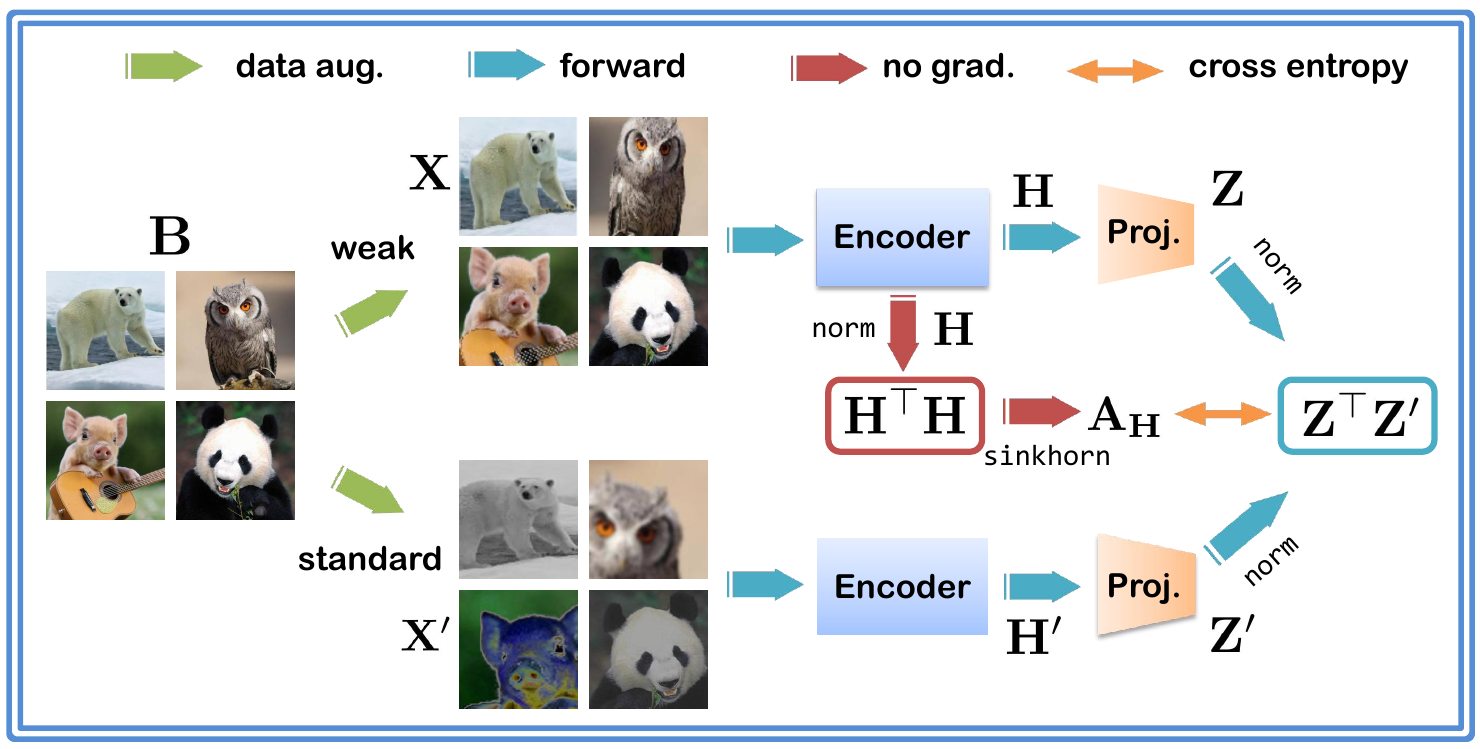}}
    \vspace{-0.1in}
    \caption{The framework of \textbf{Re}presentation \textbf{S}elf-\textbf{A}ssignment (ReSA). Here, \textit{no grad.} denotes that the operation does not involve gradient propagation, \textit{norm} signifies that each sample is $L_2$-normalized to compute cosine similarities, and \textit{sinkhorn} refers to the Sinkhorn-Knopp algorithm used for clustering assignment. }
    \label{fig:resa}
    \end{center}
     \vspace{-0.4in}
    \end{figure}
    
\vspace{-0.1in}
\section{Leverage Clustering Properties for Positive-Feedback Learning}
\vspace{-0.05in}
Based on above analyses, we design a novel positive-feedback SSL method, which derives \textbf{Re}presentation \textbf{S}elf-\textbf{A}ssignment (ReSA) to guide the loss function among \Term{embeddings} $\Z{}$ and $\Z{}'$. See Figure~\ref{fig:resa} for the clear framework.
\vspace{-0.25in}
\subsection{Online Self-Clustering} 
Following notations in Section~\ref{para:jea}, we apply the \Term{encoding} $\MH{}$ as the representation to perform clustering. Unlike previous approaches, e.g. SwAV~\cite{2020_NIPS_Caron} and DINOv2~\cite{2023_arxiv_Oquab} employing learnable prototypes to map features into the clustering space, we treat samples in $\MH{}=[\h{1},\dots,\h{m}]$ simultaneously as points to be clustered and as \textbf{anchors}. In details, we first calculate the cosine self-similarity matrix by $\MS{\MH{}}=\MH{}^\top\MH{}$, where samples in $\MH{}$ are $L_2-$normalized as $\h{i}/{\| \h{i} \|_2}, \forall i$. Then the online clustering assignment $\MA{\MH{}}$ is computed upon $\MS{\MH{}}$ using the iterative Sinkhorn-Knopp~\cite{2013_NIPS_Cuturi} as shown in Algorithm~\ref{alg:sinkhorn}.

\vspace{-0.1in}
\begin{algorithm}[ht]
\caption{Sinkhorn-Knopp Algorithm}
\label{alg:sinkhorn}
\begin{algorithmic}
\REQUIRE Cosine self-similarity matrix $ \MS{\MH{}} \in \mathbb{R}^{m \times m} $, regularization parameter $ \epsilon > 0 $, all-ones column vector $\mathbf{1}_m$, iteration count $ T $, Hadamard product $\circ$

\ENSURE Doubly stochastic matrix $\MA{\MH{}}$

\STATE Initialize $ \mathbf{Q} \gets \frac{\exp(\MS{\MH{}} / \epsilon)^\top}{\sum_{i,j} \exp(\MS{\MH{}} / \epsilon)} $
\STATE Initialize marginals: $ \mathbf{c} \gets \frac{1}{m} \mathbf{1}_m $

\FOR{$ t = 1 $ to $ T $} 
    \STATE Compute row sums: $ \mathbf{u} \gets \mathbf{Q} \mathbf{1}_m $
    \STATE Normalize rows: $ \mathbf{Q} \gets \mathbf{Q} \circ \left(\frac{\mathbf{c}}{\mathbf{u}}\right) \mathbf{1}_m^\top $
    \STATE Compute column sums: $ \mathbf{v} \gets \mathbf{Q}^\top \mathbf{1}_m $
    \STATE Normalize columns: $ \mathbf{Q} \gets \mathbf{Q} \circ \mathbf{1}_m \left(\frac{\mathbf{c}}{\mathbf{v}}\right)^\top $
\ENDFOR

\STATE Normalize columns again: $ \mathbf{Q} \gets \mathbf{Q} \circ \mathbf{1}_m \left(\frac{1}{\mathbf{Q}^\top \mathbf{1}_m}\right)^\top $

\STATE Return $\MA{\MH{}} \gets \mathbf{Q}^\top $
\end{algorithmic}
\end{algorithm}
\vspace{-0.1in}
We follow SwAV which uses only 3 iterations and sets the regularization parameter $ \epsilon =0.05 $. This algorithm does \textbf{not} involve gradient propagation, enabling it to be efficiently implemented on GPUs~\cite{2020_NIPS_Caron}. After obtaining the doubly stochastic matrix $\MA{\MH{}}$ as the assignment, it can naturally be utilized to guide relationship between the \Term{embeddings} $\Z{}$ and $\Z{}'$. Specially, we use the cross-entropy loss to promote the learning process:

%$$
%\ell = -\frac{1}{2m} \left( %\sum_{i,j} \MA{\MH{i,j}} \cdot \log %P_{i,j} + \sum_{i,j} \MA{\MH{j,i}} %\cdot \log P^T_{j,i} \right)
%$$
\vspace{-0.1in}
 \begin{equation}
 \label{eqn:resa}
 \begin{aligned}
   \ell_{\text{ReSA}} = -\frac{1}{2m} \bigg( &\sum_{i,j} \MA{\MH{}} \circ \log \mathcal{D}(\Z{}^\top\Z{}') + \\
   &   \sum_{i,j} \MA{\MH{}}^\top \circ \log \mathcal{D}(\Z{}'^\top\Z{}) \bigg)
    \end{aligned}
\end{equation}

\vspace{-0.02in}
where $\mathcal{D}(\Z{}^\top\Z{}') = \frac{\exp(\Z{}^\top\Z{}'/\tau)}{\exp(\Z{}^\top\Z{}'/\tau)\mathbf{1}_m}$ and $\mathcal{D}(\Z{}'^\top\Z{})$ are probability distributions derived through the softmax function. $\tau$ is a scalar temperature hyperparameter, $\circ$ stands for Hadamard product, and $\mathbf{1}_m$ is the all-ones column vector.

\paragraph{Comparison to SwAV.} As a pioneering SSL method based on online clustering, SwAV~\cite{2020_NIPS_Caron} employs a `swapped' prediction mechanism (which is also adopted by DINOv2~\cite{2023_arxiv_Oquab}), where the cluster assignment of one view is predicted from the \Term{embedding} of another view. This is achieved by minimizing the following objective:
\vspace{-0.15in}
 \begin{equation}
        \label{eqn:swav}
    \begin{aligned}
   \ell_{\text{SwAV}} = -\frac{1}{2m} \bigg(&\sum_{i,j} \mathbf{Q}' \circ \log \mathcal{D}(\Z{}^\top\mathbf{C}) +
    \\
    &\sum_{i,j} \mathbf{Q} \circ \log \mathcal{D}(\Z{}'^\top \mathbf{C}) \bigg)
    \end{aligned}
\end{equation}

\vspace{-0.15in}
where $\mathbf{C} \in \mathbb{R}^{d_g\times K}$ is the prototype matrix learned by back-propagation, and $\mathbf{Q}=\text{sinkhorn}(\Z{}^\top\mathbf{C})$ is the cluster assignment using Sinkhorn-Knopp algorithm. The key differences and advantages of our ReSA compared to SwAV (and DINOv2) can be summarized as follows:
(1) ReSA computes clustering assignments on \Term{encoding} with high-quality clustering properties, whereas SwAV performs it on the less stable \Term{embedding}.
(2) SwAV requires learnable prototypes, which often necessitate complex design strategies, such as freezing prototypes during the early stages of training and use a large number of prototypes $K$ to ensure stability, whereas ReSA directly extracts clustering information from the representations.
(3) SwAV executes the Sinkhorn-Knopp algorithm multiple times, corresponding to the number of global augmented views. In contrast, ReSA only requires a single execution of this algorithm regardless of the number of augmented views. This highlights the efficiency of ReSA, particularly under multi-crop scenarios.
\vspace{-0.1in}
\paragraph{Comparison to InfoNCE.} As a well-known contrastive loss function, InfoNCE~\cite{2018_arxiv_Oord} aims to maximize the similarity between positive pairs while minimizing the similarity between the negative pairs, thereby approximating the optimization of mutual information as follows:
 \begin{equation}
        \label{eqn:infonce}
   \ell_{\text{InfoNCE}} = -\frac{1}{2m} \bigg( \sum_{i,i} \log \mathcal{D}(\Z{}^\top\Z{}')  + \sum_{i,i} \log \mathcal{D}(\Z{}'^\top\Z{}) \bigg)
\end{equation}

It is evident that when $\MA{\MH{}}$ equals the identity matrix, ReSA and InfoNCE are entirely equivalent. In other words, ReSA guides the relationship among \Term{embeddings} through assignments obtained via self-clustering, whereas InfoNCE employs the identity matrix as a hard matching target, strictly enforcing the maximization of distances between all negative pairs, which may inadvertently push samples belonging to the same category further apart during training, thereby disrupting the underlying semantic cluster structure~\cite{2021_cvpr_wang_feng,2024_TPAMI_Huang}.
\vspace{-0.05in}
\subsection{A Gradient Perspective on ReSA}
Many works have conducted in-depth theoretical studies on InfoNCE, such as the alignment and uniformity properties observed by~\citet{2020_ICML_Wang}, the hardness-aware property discovered by~\citet{2021_cvpr_wang_feng} through gradient analysis, and the probabilistic model of InfoNCE derived by~\citet{bizeul2024probabilistic} from the perspective of mutual information. Since the motivation for our proposed ReSA does not have a direct connection with the evidence lower bound of mutual information, in this section, we provide an intuitive gradient analysis to further understand the optimization mechanism of ReSA.

Given the $L_2$-normalized embedding vectors $\Z{}=[\z{1},\dots,\z{m}]$ and $\Z{}'=[\z{1}',\dots,\z{m}']$, the InfoNCE formula on $\z{i}$ can be writen as (omitting the symmetric terms):
 \begin{equation}
    \label{eqn:infoNCE_2}
   \ell_{\text{InfoNCE}}(\z{i}) = - \log \left( \frac{\exp(s_{i,i} / \tau)}{\sum_{k=1}^m \exp(s_{i,k} / \tau)} \right)
\end{equation}
where $s_{i,j}=\z{i}^\top\z{j}', \forall i,j$. Defining the probability $P_{i,j} = \frac{\exp(s_{i,j} / \tau)}{\sum_{k=1}^m \exp(s_{i,k} / \tau)}$, the gradients of InfoNCE with respect to the positive similarity $s_{i,i}$ and the negative similarity $s_{i,j}$ ($i\neq j$) are formulated as~\cite{2021_cvpr_wang_feng}:

 \begin{equation}
    \label{eqn:info_g}
   \frac{\partial \ell_{\text{InfoNCE}}(\z{i})}{\partial s_{i,i}} = - \frac{1}{\tau} \sum_{k \neq i} P_{i,k},\quad 
   \frac{\partial \ell_{\text{InfoNCE}}(\z{i})}{\partial s_{i,j}} = \frac{1}{\tau}P_{i,j}
\end{equation}

Similarly, our ReSA formula on $\z{i}$ can be writen as:
 \begin{equation}
    \label{eqn:resa_2}
   \ell_{\text{ReSA}}(\z{i}) = - \sum_{j=1}^m {\mathbf{A}_\mathbf{H}}^{(i,j)} \log \left( \frac{\exp(s_{i,j} / \tau)}{\sum_{k=1}^m \exp(s_{i,k} / \tau)} \right)
\end{equation}
By contrast, the gradient of ReSA with respect to the similarity \(s_{i,j}\) for any pair of samples (\(\forall i,j\)) takes exactly the same analytical form:
 \begin{equation}
\frac{\partial \ell_{\text{ReSA}}(\z{i})}{\partial s_{i,j}} = \frac{1}{\tau} (P_{i,j} - {\mathbf{A}_\mathbf{H}}^{(i,j)} )
\end{equation}
Based on the gradient analysis above, we know that InfoNCE explicitly distinguishes between the gradient forms of positive and negative similarities. This restrictive mechanism naturally leads to harmful gradient updates for negative sample pairs within the same class. In contrast, ReSA eliminates the distinction between positive and negative samples and adapts to optimize the similarity of all sample pairs by leveraging self-clustering of the encodings, thereby addressing a key challenge in contrastive learning.

\subsection{Impact of Image Augmentation on ReSA}
\label{sec:aug}
\vspace{-0.05in}
Having introduced the learning process of ReSA, it is essential to consider another critical aspect of SSL: image augmentation, which has long been acknowledged as a key factor in enhancing the performance of self-supervised methods~\cite{2020_arxiv_Chen,2020_NIPS_Grill}. Standard practices involve employing a variety of complex transformations with random probabilities, such as \textit{ResizedCrop}, \textit{ColorJitter}, \textit{Grayscale}, \textit{GaussianBlur}, and \textit{HorizontalFlip}, to increase the task's complexity and improve the robustness of the learned representations. However, for clustering-based SSL methods, overly aggressive augmentations can distort the original image information, making it harder for the model to discern meaningful patterns~\cite{2021_NIPS_Zheng}, which may result in incorrect clustering assignments. To address this, we systematically evaluate the effect of each transformation technique on ReSA's clustering performance during training and its linear evaluation accuracy.

\begin{figure}[t]
\vspace{-0.02in}
\begin{center}
\centerline{\includegraphics[width=8.5cm]{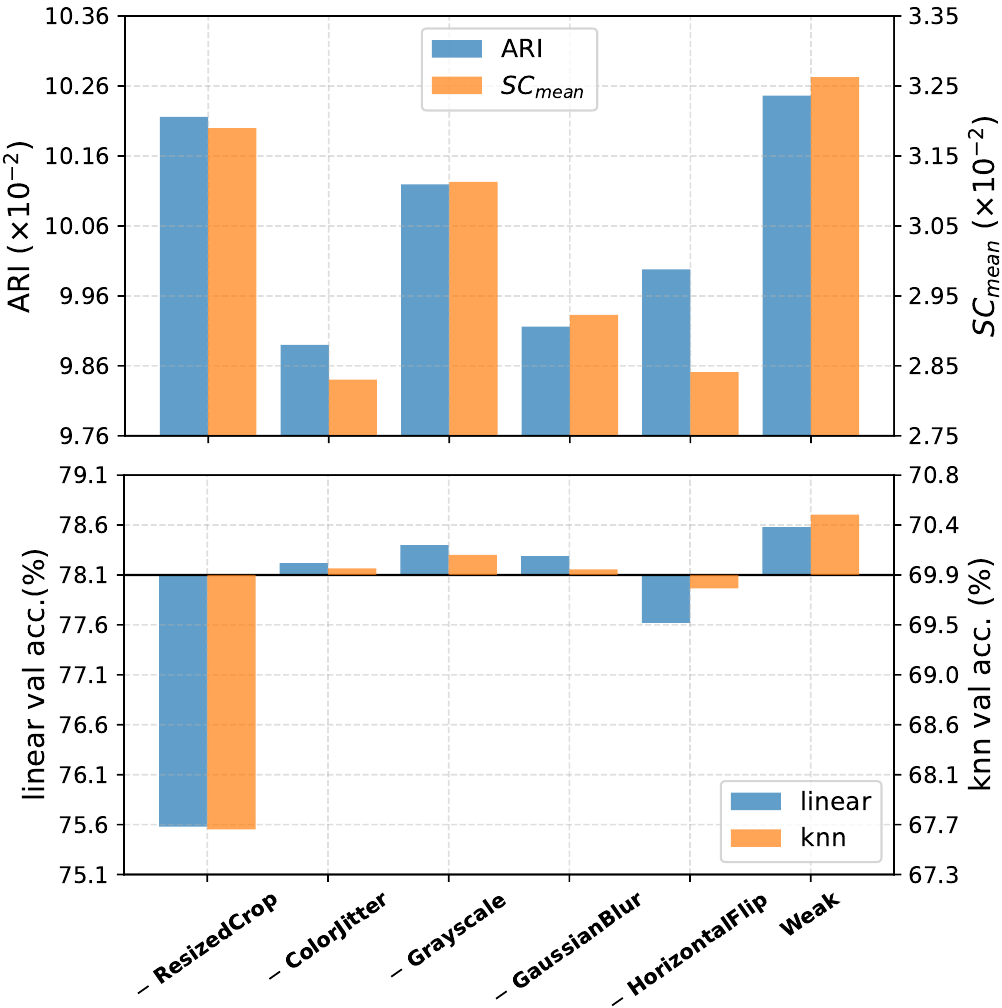}}
\vspace{-0.2in}
\caption{Investigate the impact of image augmentation on ReSA. The experiments are conducted employing a ResNet-18 encoder pretrained on ImageNet-100 for 200 epochs. The starting positions of bars represent results of the standard augmentation. The symbols `$-$' on the $x$-axis indicate the removal of a specific transformation from the standard augmentation. `Weak' denotes the weak augmentation that includes only \textit{ResizedCrop} and \textit{HorizontalFlip}. }
\label{fig:in100_aug}
\end{center}
\vspace{-0.46in}
\end{figure}

\begin{table*}[t]
 \vspace{-0.15in}
    \centering
    \caption{Classification top-1 accuracies of a \textit{linear} and a \textit{k-nearest neighbors} ($k=5$) classifier for different loss functions and datasets. The table is mostly inherited from solo-learn~\cite{solo_learn_victor}. All methods are based on ResNet-18 with two augmented views generated from per sample and are trained for 1000-epoch on CIFAR-10/100 with a batch size of 256 and 400-epoch on ImageNet-100 with a batch size of 128. The bold values indicate the best performance, and the underlined values represent the second highest accuracy.}
     \vspace{0.05in}
    \footnotesize
    \setlength{\tabcolsep}{10pt}  % Increase column spacing for readability
    \renewcommand{\arraystretch}{1}  % Increase row height for better spacing
    \begin{tabular}{l|cc|cc|cc}
        \toprule
        \multirow{2}{*}{Method} & \multicolumn{2}{c|}{CIFAR-10} & \multicolumn{2}{c|}{CIFAR-100} & \multicolumn{2}{c}{ImageNet-100} \\
        & \textit{linear} & \textit{k-nn} & \textit{linear} & \textit{k-nn} & \textit{linear} & \textit{k-nn} \\
        \arrayrulecolor{gray}
        \midrule
        SimCLR~\cite{2020_ICML_Chen} & 90.74 & 85.13 & 65.78 & 53.19 & 77.64 & 65.78 \\
        BYOL~\cite{2020_NIPS_Grill} & 92.58 & 87.40 & 70.46 & 56.46 & 80.32 & 68.94 \\
        SwAV~\cite{2020_NIPS_Caron} & 89.17 & 84.18 & 64.88 & 53.32 & 74.28 & 63.84 \\
        SimSiam~\cite{2021_CVPR_Chen} & 90.51 & 86.82 & 66.04 & 55.79 & 78.72 & 67.92 \\
        MoCoV3~\cite{2021_mocov3_Chen} & \underline{93.10} & 89.47 & 68.83 & 58.23 & 80.36 & 72.76 \\
        W-MSE~\cite{2021_ICML_Ermolov} & 91.55 & 89.69 & 66.10 & 56.69 & 76.23 & 67.72 \\
        DINO~\cite{2021_ICCV_Caron} & 89.52 & 86.13 & 66.76 & 56.24 & 74.92 & 64.30 \\
        Barlow Twins~\cite{2021_NIPS_Zbontar} & 92.10 & 88.09 & \underline{70.90} & 59.40 & 80.16 & 72.14 \\
        VICReg~\cite{2022_ICLR_Adrien} & 92.07 & 87.38 & 68.54 & 56.32 & 79.40 & 71.94 \\
        CW-RGP~\cite{2022_NIPS_Weng} & 92.03 & 89.67 & 67.78 & 58.24 & 76.96 & 68.46 \\
        INTL~\cite{2024_ICLR_Weng} & 92.60 & \underline{90.03} & 70.88 & \underline{61.90} & \underline{81.68} & \underline{73.46} \\
        \midrule
        \textbf{ReSA (ours)} & \textbf{93.53} & \textbf{93.02} & \textbf{72.21} & \textbf{66.83} & \textbf{82.24} & \textbf{74.56} \\
        \arrayrulecolor{black}
        \bottomrule
    \end{tabular}
    \label{tab:small_medium_baseline}
    \vspace{-0.2in}
    \end{table*}
    
\vspace{-0.05in}
Since ReSA only requires extracting clustering information from the \Term{encodings} of \textbf{one single view}, we only need to adjust the image augmentation for that specific view, while the standard augmentation setting can be applied to the other view(s). Subsequently, we conduct experiments using the high-resolution ImageNet-100~\cite{2020_ECCV_Tian} dataset, and the results are presented in Figure~\ref{fig:in100_aug}. It is evident that removing any single transformation improves the clustering performance of the representations, with \textit{ResizedCrop} (replaced with fixed \textit{CenterCrop}) having the most significant impact. However, its removal leads to a substantial decline in representation quality, indicating the critical role of \textit{ResizedCrop} in learning invariance. The removal of \textit{ColorJitter}, \textit{Grayscale}, or \textit{GaussianBlur} each results in improvements across various metrics, whereas removing \textit{HorizontalFlip} causes a slight drop in val accuracies. Based on these findings, we design a weak augmentation for ReSA, which includes only \textit{ResizedCrop} and \textit{HorizontalFlip}, and discover that this design not only significantly enhances the clustering performance of the representations but also improves their overall quality. We note that these findings align with the results observed in ReSSL~\cite{2021_NIPS_Zheng}, where weak augmentation enables the model to better capture the relationships among samples.

In summary, we have completed the introduction of the ReSA framework as shown in Figure~\ref{fig:resa}, which leverages clustering information extracted from the \Term{encoding} to guide the design of the loss function, thereby achieving positive-feedback self-supervised learning.

\vspace{-0.05in}
\section{Experiments on Standard SSL Benchmark} 
\label{sec:benchmark_experiments}
\vspace{-0.05in}
In this section, we conduct extensive experiments on standard SSL benchmarks to evaluate the effectiveness of ReSA. We perform pretraining from scratch on a variety of datasets, including CIFAR-10/100, ImageNet-100, and ImageNet~\cite{2009_ImageNet}, utilizing diverse encoders such as ConvNets and the ViT. Furthermore, we compare the performance of ReSA with state-of-the-art SSL methods across a range of downstream tasks, e.g. linear probe evaluation and transfer learning. The full PyTorch-style algorithm as well as details of implementation is provided in Appendix~\ref{appendix:implementation}.
\begin{table}[t]
\vspace{-0.1in}
    \centering
    \caption{ImageNet classification top-1 accuracy of a \textit{linear} classifier based on ResNet-50 encoder. All methods are pretrained with \textbf{two $224^2$ augmented views} generated from per sample. Given that one of the objectives of SSL methods is to achieve high performance with small batch sizes~\cite{2020_arxiv_Chen,2021_CVPR_Chen}, it's worth noting that our ReSA can also perform effectively when trained with a small batch size of 256.}
     \vspace{0.1in}
    \footnotesize
    \setlength{\tabcolsep}{8pt}  % Increase column spacing for readability
    \renewcommand{\arraystretch}{0.92}  % Increase row height for better spacing
    \begin{tabular}{lcccc}
        \toprule
        \multirow{2}{*}{Method} & \multirow{2}{*}{batch size} & \multicolumn{3}{c}{pretrained epochs} \\
        & &  100  & 200 & 800  \\
        \midrule
        \multirow{2}{*}{SimCLR} &256& 57.5 & 62.0 & 66.5 \\
        & 4096 &  66.5 & 68.3 & 70.4 \\
        \arrayrulecolor{gray}
        \midrule
        \multirow{2}{*}{SwAV} &256 &  65.5 & 67.7 & - \\
        & 4096 & 66.5 & 69.1 & 71.8\\
        \midrule
        \multirow{2}{*}{MoCoV3} &1024 &  67.4 & 71.0 & 72.4 \\
        & 4096 & 68.9 & - & 73.8 \\
        \midrule
        BYOL & 4096 & 66.5 & 70.6 & 74.3 \\
        Barlow Twins & 2048 &  67.7 & 70.2 & 73.2 \\
        VICReg & 2048 & 68.6 & 70.8  & 73.1  \\
        \midrule
        \multirow{2}{*}{SimSiam} & 256 &  68.1 & 70.0 & 71.3\\
       & 1024 & 68.0 & 69.9 & 71.1 \\
       \midrule
       \multirow{2}{*}{MEC} & 256  & 70.1 & - & -\\
       & 1024 & 70.6 & 71.9 & 74.0 \\
       \midrule
       \multirow{2}{*}{INTL} & 256 &  69.5 & 71.1 & 73.1\\
       & 1024 & 69.7 & 71.2 & 73.3\\
        \midrule
        \multirow{2}{*}{\textbf{ReSA (ours)}} & 256 & \textbf{71.9} & \textbf{73.4} & - \\
        & 1024 & \textbf{71.3} & \textbf{73.8} & \textbf{75.2} \\
        \arrayrulecolor{black}
        \bottomrule
    \end{tabular}
    \label{tab:imagenet_epochs}
    \vspace{-0.3in}
\end{table}

\vspace{-0.05in}
\subsection{Evaluation for Classification}

\paragraph{Evaluation on small and medium size datasets.} 
Following the benchmark in solo-learn~\cite{solo_learn_victor}, we first perform pretraining and classification evaluations on CIFAR-10/CIFAR-100, and ImageNet-100, strictly adhering to the same experimental settings as other methods without introducing any additional tricks. The results in Table~\ref{tab:small_medium_baseline} reveal that ReSA consistently outperforms state-of-the-art methods, even those with carefully optimized parameters, across all datasets. Particularly noteworthy is ReSA’s performance in \textit{k-nearest neighbors} classification, surpassing other methods with absolute accuracy improvements of approximately $3\%$ to $8\%$ on CIFAR-10 and $5\%$ to $13\%$ on CIFAR-100. These findings highlight that ReSA captures representations with superior clustering structures.
\begin{table}[t]
    \vspace{-0.1in}
    \centering
    \caption{ImageNet classification top-1 accuracy of a \textit{linear} and a \textit{k-nearest neighbors} ($k=20$) classifier based on a standard ViT-S/16 encoder. All models are pretrained for 300-epoch with two $224^2$ views.}
    \vspace{0.05in}
    \footnotesize
    \setlength{\tabcolsep}{5pt}  % Increase column spacing for readability
    \renewcommand{\arraystretch}{1}  % Increase row height for better spacing
    \begin{tabular}{ccccccc}
        \toprule
        Classifier & BYOL & SwAV & MoCoV3 & DINO & \textbf{ReSA} \\
       \arrayrulecolor{gray}
        \midrule
     \textit{linear} & 71.4 & 68.5 & 72.5 & 72.5 & \textbf{72.7} \\
     \textit{k-nn} & 66.6 & 60.5 & 67.7 & 67.9 & \textbf{68.3} \\
    \arrayrulecolor{black}
    \bottomrule
    \end{tabular}
    \label{tab:vit}
    \vspace{-0.15in}
\end{table}

\begin{table}[t]
    \centering
    \caption{Comparison of Computational overhead among various SSL methods. For fairness, we set the batch size to 1024 with two $224^2$ augmented views pretraining on ImageNet, and perform all measurements including peak memory (GB per GPU) and training time (hours per epoch) on the same environment and machine equipped with 8 A100-PCIE-40GB GPUs using 32 dataloading workers under mixed-precision. }
    \vspace{0.05in}
    \footnotesize
    \setlength{\tabcolsep}{5pt}  % Increase column spacing for readability
    \renewcommand{\arraystretch}{1}  % Increase row height for better spacing
    \begin{tabular}{ccccccc}
        \toprule
        Method & encoder & memory & time \\
       \arrayrulecolor{gray}
        \midrule
     \multirow{2}{*}{SwAV} & ResNet-50 &   \textbf{13.7} &  0.19 &\\
      
     & ViT-S/16 &  \textbf{14.8} &  0.17  &\\
     \arrayrulecolor{gray}
        \midrule
     \multirow{2}{*}{MoCoV3} & ResNet-50 &   14.6 &   0.19  &\\
     
     & ViT-S/16 &  15.5 &   0.22 &\\
     \arrayrulecolor{gray}
        \midrule
      \multirow{2}{*}{DINO} & ResNet-50 &   15.5 &   0.20 &\\
   
     & ViT-S/16 &  16.4 &   0.21 &\\
     \arrayrulecolor{gray}
        \midrule
     \multirow{2}{*}{\textbf{ReSA (ours)}} & ResNet-50 &  14.6 &  \textbf{ 0.16} &\\
 
     & ViT-S/16 &  15.6 &  \textbf{ 0.12}  &\\
    \arrayrulecolor{black}
    \bottomrule
    \end{tabular}
    \label{tab:overhead}
    \vspace{-0.15in}
\end{table}

\vspace{-0.1in}
\paragraph{Evaluation on ImageNet.}  Following the ImageNet evaluation protocol commonly used by SSL methods, we pretrain ResNet-50 encoders with ReSA for varying numbers of epochs. As shown in Table~\ref{tab:imagenet_epochs}, ReSA consistently outperforms other methods on the large-scale ImageNet dataset. Remarkably, after only 100-epoch training, ReSA surpasses the performance of SimCLR, SwAV, and SimSiam trained for 800 epochs. With 200 epochs, ReSA exceeds state-of-the-art methods such as MoCoV3, Barlow Twins, VICReg, and INTL, all trained for 800 epochs. When extended to 800 epochs, ReSA achieves a linear classification accuracy of $75.2\%$, a level that methods like SwAV and DINO only reach by employing the multi-crop~\cite{2020_NIPS_Caron} trick. These results underscore ReSA's exceptional potential for training on large-scale datasets. Additionally in Table~\ref{tab:vit}, we conduct preliminary evaluations of ReSA's training capability on the Vision Transformer, using a standard ViT-S/16, which has a comparable number of parameters to ResNet-50. We do not incorporate extensive training tricks, yet ReSA still outperform DINO in both \textit{linear} and \textit{k-nn} classification.

\vspace{-0.1in}
\subsection{Analysis on Computational Overhead}
\vspace{-0.05in}
In Table~\ref{tab:overhead}, we provide a fair comparison of the training costs among ReSA and several SSL methods. The results show that ReSA has memory consumption comparable to MoCoV3 but achieves faster training speeds, especially on ViTs, where it is nearly twice as fast. This improvement is attributed to ReSA’s simpler image augmentation settings and the removal of batch normalization (BN) from the projector and predictor MLPs. Additionally, ReSA outperforms DINO in both memory usage and training time. This advantage stems from DINO's reliance on an extremely high-dimensional prototype (e.g., $\text{output dimension} = 65536$), which significantly impacts training efficiency. Finally, while SwAV exhibits the smallest memory footprint among the methods due to its absence of momentum networks, its training speed remains slower than ReSA. This is because SwAV also requires a higher-dimensional prototype and performs the Sinkhorn-Knopp algorithm twice per iteration.

\begin{table}[t]
    \vspace{-0.1in}
    \centering
        \caption{Transfer Learning to COCO detection and instance segmentation. All competitive methods are based on ResNet-50 with 200-epoch pretraining on ImageNet. We follow MoCo~\cite{2020_CVPR_He} to apply Mask R-CNN (1 $\times$ schedule) fine-tuned in COCO 2017 train, evaluated in COCO 2017 val.}
         \vspace{0.05in}
        \footnotesize
        \setlength{\tabcolsep}{5.5pt}  % Increase column spacing for readability
        \renewcommand{\arraystretch}{1}  % Increase
            \begin{tabular}{lccccccc}
                \toprule
                \multirow{2}{*}{Method} & 
                \multicolumn{3}{c}{COCO detection} &
                \multicolumn{3}{c}{COCO instance seg.} \\
                &AP$_{50}$&AP&AP$_{75}$ &AP$_{50}$&AP&AP$_{75}$ \\
                 \midrule
                Scratch &44.0&	26.4&	27.8&	46.9&	29.3&	30.8 \\
                Supervised &58.2&	38.2&	41.2&	54.7&	33.3&	35.2 \\
                \arrayrulecolor{gray}
                 \midrule
                SimCLR &57.7&37.9&	40.9&	54.6&	33.3&	35.3 \\	
                MoCoV2 &	58.8&	39.2&	42.5&	55.5&	34.3&	36.6 \\
                BYOL &	57.8&	37.9&	40.9&	54.3&	33.2&	35.0 \\
                SwAV & 57.6& 37.6&40.3&54.2&33.1&35.1 \\
                SimSiam & 59.3 &39.2 &42.1 &56.0 &34.4 &36.7 \\
                %W-MSE &
                %-60.1&39.2&	42.8&56.8&34.8&36.7 \\
                Barlow Twins &	59.0&39.2&42.5&	56.0&	34.3&	36.5 \\
                MEC  &59.8 &39.8 &43.2 &56.3 &34.7 &36.8 \\
                INTL &60.9&	40.7&43.7&57.3&	35.4 &	37.6 \\
                \midrule
                \textbf{ReSA (ours)} &\textbf{61.1}&	\textbf{41.0}&\textbf{44.3}&\textbf{57.7}& \textbf{35.7}&	\textbf{38.4} \\
                \arrayrulecolor{black}
                \bottomrule
            \end{tabular}
    \label{tab:transfer}
     \vspace{-0.3in}
    \end{table}

 \vspace{-0.1in}
\subsection{Transfer to Downstream Tasks}
\vspace{-0.05in}
To evaluate the quality of representations learned by ReSA, we transfer our pretrained model to downstream tasks, including COCO~\cite{2014_ECCV_COCO} object detection and instance segmentation. For these tasks, we adopt the baseline codebase from MoCo~\cite{2020_CVPR_He}. Most results reported in Table~\ref{tab:transfer} are inherited from SimSiam paper~\cite{2021_CVPR_Chen}. Notably, ReSA also achieves better performance compared to other methods on both tasks, highlighting its strong potential for downstream applications.

    \begin{figure}[t]
    %\vskip 0.2in
    \begin{center}
    \centerline{\includegraphics[width=8.2cm]{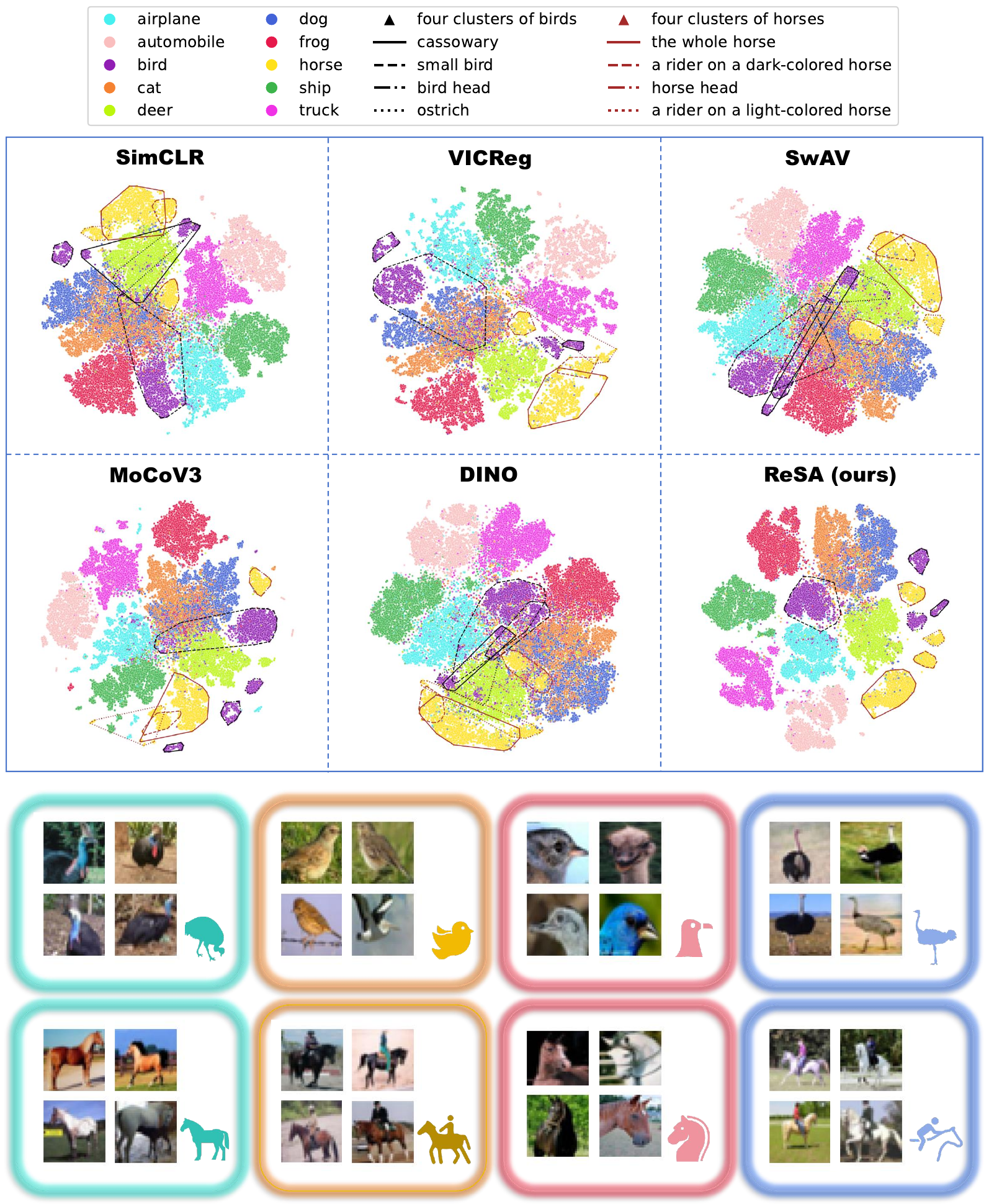}}
    \caption{T-SNE visualization of SSL representations on CIFAR-10. All methods are pretrained for 1000 epochs on CIFAR-10 using ResNet-18, with \Term{encodings} utilized as representations to visualize all training data. For the multiple centroids observed in the bird and horse categories, we enclose points of each subclass with convex polygons and display the corresponding images.}
    \label{fig:cifar10_tsne}
    \end{center}
    \vspace{-0.3in}
    \end{figure}

\begin{table*}[t]
    \vspace{-0.1in}
    \centering
        \caption{Transfer learning to fine-grained datasets based on ResNet-50 pretrained on ImageNet. We employ a \textit{k-nearest neighbors} classifier ($k=5, 10, 20$), without requiring additional training or parameter tuning. The model weights for all other methods are sourced directly from their respective official codebases. $\dag$ indicates that these methods employ the multi-crop trick, i.e. generating two $224^2$ views and six $96^2$ views for each image, which can enhance performance but comes at the cost of additional computational overhead.}
         \vspace{0.05in}
        \footnotesize
        \setlength{\tabcolsep}{4pt}  % Increase column spacing for readability
        \renewcommand{\arraystretch}{1.1}  % Increase
        \begin{tabular}{c|c|ccc|ccc|ccc|ccc|ccc}
        \toprule
            \multirow{2}{*}{Method} &  \multicolumn{1}{c|}{pretrained} &
            \multicolumn{3}{c|}{ImageNet-1K}  &\multicolumn{3}{c|}{CUB-200-2011}  & \multicolumn{3}{c|}{Pets-37} & \multicolumn{3}{c|}{Food-101} & \multicolumn{3}{c}{Flowers-102} \\
            & epochs & $5$ & $10$ & $20$ & $5$ & $10$ & $20$ & $5$ & $10$ & $20$ & $5$ & $10$ & $20$ & $5$ & $10$ & $20$ \\ \midrule
            MoCoV3 & 1000 & 67.9 & 68.9 & 68.9  & 46.8 & 48.8 & 50.4   & 85.4 & 86.5 & 86.5 & 56.3 & 58.6 &59.7  & 83.4 & 81.6 & 80.9\\
            VICReg & 1000 & 64.3 & 65.2 &65.6  & 33.4 & 35.4 & 36.3  & 81.5 & 82.0 & 82.3 & 56.9 & 59.6 & 61.0  & 83.4 & 83.2 & 82.6\\
            INTL & 800 & 63.6 & 64.8  & 65.1 & 26.7 & 28.0  & 29.4  & 78.4 & 79.5 & 79.7  & 55.6 & 58.1 & 59.2 & 78.8 & 77.6 & 77.2\\
            \textbf{ReSA (ours)} & 800 & \textbf{69.2} & \textbf{69.9} & \textbf{69.9} & \textbf{56.5} & \textbf{58.5} & \textbf{59.9}   & \textbf{85.8} & \textbf{87.2} & \textbf{87.5} & \textbf{58.3} & \textbf{60.4} & \textbf{61.3} & \textbf{84.4} & \textbf{83.6} & \textbf{83.6} \\ 
            \arrayrulecolor{gray}
            \midrule
            SwAV$^{\dag}$ & 800 & 64.3 & 65.5 & 65.7  & 26.2 & 27.3 & 28.4  & 77.2 & 77.3 & 77.1 & 54.7 & 57.4 & 58.7 & 79.3 & 79.9 & 78.4\\
            DINO$^{\dag}$ & 800 & 66.4 & 67.4 & 67.6  & 33.8 & 35.5 & 36.8    & 81.1 & 81.6 & 80.9  & 58.2 & \textbf{60.8} & \textbf{61.8} & \textbf{84.8} & \textbf{84.1} & \textbf{83.7}\\
            \arrayrulecolor{black}
            \bottomrule
        \end{tabular}
        \label{tab:fine_grained}
        \vspace{-0.1in}
\end{table*}

\section{How ReSA Shapes Better Clustering Properties?}
In this section, we utilize visualizations and additional experiments to illustrate the differences among the representations learned by ReSA and other SSL methods, as well as to investigate whether and how ReSA facilitates better clustering properties.

Firstly, we track the evolution of clustering metrics for each component during ReSA training, as shown in Figure~\ref{fig:cifar100_compare}. It can be clearly observed that while ReSA exhibits relatively slow performance improvement in the early stages of training, it significantly outperforms other methods in the later stages. Interestingly, all components of ReSA demonstrate strong clustering properties, suggesting that leveraging high-quality clustering information from the \Term{encodings} to guide the learning of \Term{embeddings} enables the projector layers to also acquire robust clustering performance.

\subsection{ReSA Excels at Fine-grained Learning}

Furthermore, in Figure~\ref{fig:cifar10_tsne}, we visualize the representation distributions learned by ReSA and other SSL methods on CIFAR-10 using T-SNE~\cite{Maaten2008VisualizingDU}. Notably, the representations learned by ReSA exhibit clear separations between different classes, whereas those learned by other methods show varying degrees of overlap, making it difficult to discern distinct boundaries.

\vspace{-0.02in}
Another intriguing observation is the presence of multiple centroids within the bird and horse categories in the representations. Upon further investigation of the samples corresponding to these centroids, we find that, unlike the neural collapse~\cite{2020_NAD_Papyan} in supervised learning (where samples of the same class collapse to a single point), SSL models are capable of capturing more fine-grained features, such as color distinctions (e.g., cassowary vs. ostrich), structural differences (e.g., whole horse vs. horse head), and the presence of multiple objects (e.g., a rider on a horse). Moreover, compared to other methods, ReSA demonstrates a superior ability to distinguish these fine-grained features. To further substantiate this, we transfer models pretrained on ImageNet to fine-grained datasets for evaluation. As shown in Table~\ref{tab:fine_grained}, ReSA consistently outperforms other SSL methods on fine-grained datasets, with particularly considerable improvements observed on the CUB-200-2011.

    \begin{figure}[t]
    %\vskip 0.2in
    \vspace{-0.05in}
    \begin{center}
    \centerline{\includegraphics[width=8.2cm]{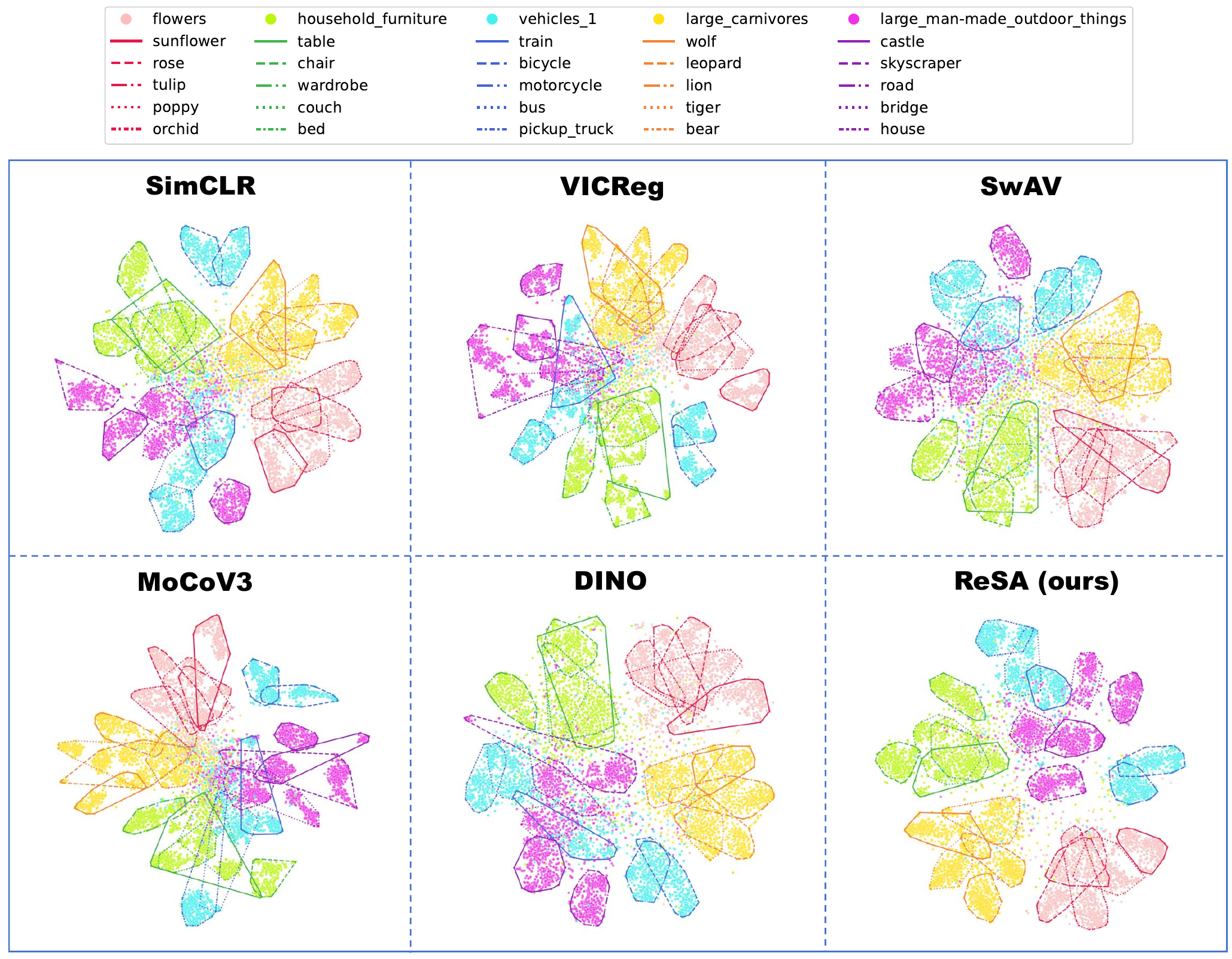}}
    \vspace{-0.1in}
    \caption{T-SNE visualization of SSL representations on CIFAR-100. We enclose points of each subclass with convex polygons.}
    \label{fig:cifar100_tnse}
    \end{center}
    \vspace{-0.4in}
    %\vskip -0.2in
    \end{figure}

\subsection{ReSA also Stands Out in Coarse-grained Representations}

Finally, we explore ReSA's performance at the coarse-grained level. Specifically, we select CIFAR-100 for visualization, as its 100 classes can be grouped into 20 coarse-grained superclasses. For clarity, we randomly selected 5 superclasses for T-SNE visualization, as shown in Figure~\ref{fig:cifar100_tnse}. It is evident that all other methods exhibit dense overlap on the CIFAR-100 dataset, resulting in numerous indistinguishable outliers. In contrast, our ReSA effectively identifies these hard samples, clustering them correctly within their respective groups. We believe this capability of ReSA is a key factor behind its superior accuracy with the \textit{k-nn} classifier, substantially exceeding other SSL methods. Furthermore, as shown in Table~\ref{tab:corse_grained}, we evaluate the performance of various SSL models using coarse-grained labels and observe that ReSA consistently achieves much higher accuracies than its counterparts. These experimental results confirm that ReSA also demonstrates exceptional performance in coarse-grained learning.

We also present ablation studies, along with a discussion of potential future research presented in Appendix~\ref{sec:ablation}.

\begin{table}[t]
     \vspace{-0.1in}
    \centering
        \caption{CIFAR-100 classification top-1 accuracy of a \textit{linear} and a \textit{k-nearest neighbors} ($k=5$) classifier based on 100 fine-grained classes and 20 coarse-grained superclasses.}
         \vspace{0.05in}
        \footnotesize
        \setlength{\tabcolsep}{7pt}  % Increase column spacing for readability
        \renewcommand{\arraystretch}{1.1}  % Increase
        \begin{tabular}{c|cc|cc}
        \toprule
            \multirow{2}{*}{Method} &  \multicolumn{2}{c|}{fine-grained}  & \multicolumn{2}{c}{coarse-grained} \\
            & \textit{linear} & \textit{k-nn}  &\textit{linear} & \textit{k-nn}\\ \midrule
            SimCLR & 65.8 & 53.2 & 72.5  & 67.2    \\
            SwAV & 64.9 & 53.3 & 70.0  & 66.3  \\
            MoCoV3 & 68.8 & 58.2 & 76.4  & 68.6 \\
            DINO & 66.8 & 56.2 & 72.9 & 70.2   \\ 
            VICReg & 68.5 & 56.3 & 74.3 & 69.9  \\
            \arrayrulecolor{gray}
        \midrule
            \textbf{ReSA (ours)} & \textbf{72.2} & \textbf{66.8} & \textbf{79.8}   & \textbf{78.8} \\ 
    \arrayrulecolor{black}
    \bottomrule
        \end{tabular}
        \label{tab:corse_grained}
    \end{table}

\section{Conclusion}
In this work, we demonstrate the feasibility of leveraging the rich clustering properties inherent in SSL models, particularly within \Term{encodings}, to enable a positive-feedback mechanism. Building upon this, we propose ReSA, which exhibits exceptional performance across a wide range of benchmarks and excels at both fine-grained and coarse-grained learning. We believe this dual capability would take a step toward addressing the long-standing challenge of reconciling the seemingly conflicting demands of fine-grained and coarse-grained visual representations within a unified framework, thereby advancing the development of large-scale visual foundation models.

\section*{Acknowledgment}
This work was supported by the  National Science and Technology Major Project (2022ZD0116306), National Natural Science Foundation of China (Grant No. 62476016 and  62441617), the Fundamental Research Funds for the Central Universities.

\section*{Impact Statement}

This paper presents work aimed at advancing the field of self-supervised learning. By improving representation learning, our methods can contribute to a wide range of applications across various domains, including computer vision, natural language processing, and beyond. While our work has potential societal implications, such as enabling more efficient use of data and reducing reliance on labeled datasets, we do not identify any specific ethical concerns or negative consequences that require particular attention at this stage.


\bibliography{reference}

\begin{thebibliography}{55}
\providecommand{\natexlab}[1]{#1}
\providecommand{\url}[1]{\texttt{#1}}
\expandafter\ifx\csname urlstyle\endcsname\relax
  \providecommand{\doi}[1]{doi: #1}\else
  \providecommand{\doi}{doi: \begingroup \urlstyle{rm}\Url}\fi

\bibitem[Asano et~al.(2019)Asano, Rupprecht, and Vedaldi]{asano2019critical}
Asano, Y.~M., Rupprecht, C., and Vedaldi, A.
\newblock A critical analysis of self-supervision, or what we can learn from a single image.
\newblock \emph{arXiv preprint arXiv:1904.13132}, 2019.

\bibitem[Assran et~al.(2022{\natexlab{a}})Assran, Balestriero, Duval, Bordes, Misra, Bojanowski, Vincent, Rabbat, and Ballas]{assran2022hidden}
Assran, M., Balestriero, R., Duval, Q., Bordes, F., Misra, I., Bojanowski, P., Vincent, P., Rabbat, M., and Ballas, N.
\newblock The hidden uniform cluster prior in self-supervised learning.
\newblock \emph{arXiv preprint arXiv:2210.07277}, 2022{\natexlab{a}}.

\bibitem[Assran et~al.(2022{\natexlab{b}})Assran, Caron, Misra, Bojanowski, Bordes, Vincent, Joulin, Rabbat, and Ballas]{2022_ECCV_Assran}
Assran, M., Caron, M., Misra, I., Bojanowski, P., Bordes, F., Vincent, P., Joulin, A., Rabbat, M., and Ballas, N.
\newblock Masked siamese networks for label-efficient learning.
\newblock In \emph{European Conference on Computer Vision}, pp.\  456--473. Springer, 2022{\natexlab{b}}.

\bibitem[Assran et~al.(2023)Assran, Duval, Misra, Bojanowski, Vincent, Rabbat, LeCun, and Ballas]{2023_cvpr_Assran}
Assran, M., Duval, Q., Misra, I., Bojanowski, P., Vincent, P., Rabbat, M., LeCun, Y., and Ballas, N.
\newblock Self-supervised learning from images with a joint-embedding predictive architecture.
\newblock In \emph{Proceedings of the IEEE/CVF Conference on Computer Vision and Pattern Recognition}, pp.\  15619--15629, 2023.

\bibitem[Bachman et~al.(2019)Bachman, Hjelm, and Buchwalter]{2019_NIPS_Bachman}
Bachman, P., Hjelm, R.~D., and Buchwalter, W.
\newblock Learning representations by maximizing mutual information across views.
\newblock In \emph{NeurIPS}, 2019.

\bibitem[Bao et~al.(2021)Bao, Dong, Piao, and Wei]{bao2021beit}
Bao, H., Dong, L., Piao, S., and Wei, F.
\newblock Beit: Bert pre-training of image transformers.
\newblock \emph{arXiv preprint arXiv:2106.08254}, 2021.

\bibitem[Bardes et~al.(2022)Bardes, Ponce, and LeCun]{2022_ICLR_Adrien}
Bardes, A., Ponce, J., and LeCun, Y.
\newblock Vicreg: Variance-invariance-covariance regularization for self-supervised learning.
\newblock In \emph{ICLR}, 2022.

\bibitem[Ben-Shaul et~al.(2023)Ben-Shaul, Shwartz-Ziv, Galanti, Dekel, and LeCun]{2023_NIPS_Ben-Shaul}
Ben-Shaul, I., Shwartz-Ziv, R., Galanti, T., Dekel, S., and LeCun, Y.
\newblock Reverse engineering self-supervised learning.
\newblock \emph{Advances in Neural Information Processing Systems}, 36:\penalty0 58324--58345, 2023.

\bibitem[Bendidi et~al.(2023)Bendidi, Bardes, Cohen, Lamiable, Bollot, and Genovesio]{bendidi2023no}
Bendidi, I., Bardes, A., Cohen, E., Lamiable, A., Bollot, G., and Genovesio, A.
\newblock No free lunch in self supervised representation learning.
\newblock \emph{arXiv preprint arXiv:2304.11718}, 2023.

\bibitem[Bizeul et~al.(2024)Bizeul, Sch{\"o}lkopf, and Allen]{bizeul2024probabilistic}
Bizeul, A., Sch{\"o}lkopf, B., and Allen, C.
\newblock A probabilistic model behind self-supervised learning.
\newblock \emph{arXiv preprint arXiv:2402.01399}, 2024.

\bibitem[Bossard et~al.(2014)Bossard, Guillaumin, and Van~Gool]{bossard2014food}
Bossard, L., Guillaumin, M., and Van~Gool, L.
\newblock Food-101--mining discriminative components with random forests.
\newblock In \emph{Computer vision--ECCV 2014: 13th European conference, zurich, Switzerland, September 6-12, 2014, proceedings, part VI 13}, pp.\  446--461. Springer, 2014.

\bibitem[Caron et~al.(2018)Caron, Bojanowski, Joulin, and Douze]{2018_ECCV_Caron}
Caron, M., Bojanowski, P., Joulin, A., and Douze, M.
\newblock Deep clustering for unsupervised learning of visual features.
\newblock In \emph{ECCV}, 2018.

\bibitem[Caron et~al.(2020)Caron, Misra, Mairal, Goyal, Bojanowski, and Joulin]{2020_NIPS_Caron}
Caron, M., Misra, I., Mairal, J., Goyal, P., Bojanowski, P., and Joulin, A.
\newblock Unsupervised learning of visual features by contrasting cluster assignments.
\newblock In \emph{NeurIPS}, 2020.

\bibitem[Caron et~al.(2021)Caron, Touvron, Misra, Jegou, Mairal, Bojanowski, and Joulin]{2021_ICCV_Caron}
Caron, M., Touvron, H., Misra, I., Jegou, H., Mairal, J., Bojanowski, P., and Joulin, A.
\newblock Emerging properties in self-supervised vision transformers.
\newblock In \emph{ICCV}, 2021.

\bibitem[Chen et~al.(2020{\natexlab{a}})Chen, Kornblith, Norouzi, and Hinton]{2020_ICML_Chen}
Chen, T., Kornblith, S., Norouzi, M., and Hinton, G.
\newblock A simple framework for contrastive learning of visual representations.
\newblock In \emph{ICML}, 2020{\natexlab{a}}.

\bibitem[Chen \& He(2021)Chen and He]{2021_CVPR_Chen}
Chen, X. and He, K.
\newblock Exploring simple siamese representation learning.
\newblock In \emph{CVPR}, 2021.

\bibitem[Chen et~al.(2020{\natexlab{b}})Chen, Fan, Girshick, and He]{2020_arxiv_Chen}
Chen, X., Fan, H., Girshick, R., and He, K.
\newblock Improved baselines with momentum contrastive learning.
\newblock \emph{arXiv preprint arXiv:2003.04297}, 2020{\natexlab{b}}.

\bibitem[Chen et~al.(2021)Chen, Xie, and He]{2021_mocov3_Chen}
Chen, X., Xie, S., and He, K.
\newblock An empirical study of training self-supervised vision transformers.
\newblock In \emph{CVPR}, 2021.

\bibitem[Cuturi(2013)]{2013_NIPS_Cuturi}
Cuturi, M.
\newblock Sinkhorn distances: Lightspeed computation of optimal transport.
\newblock \emph{Advances in neural information processing systems}, 26, 2013.

\bibitem[da~Costa et~al.(2022)da~Costa, Fini, Nabi, Sebe, and Ricci]{solo_learn_victor}
da~Costa, V. G.~T., Fini, E., Nabi, M., Sebe, N., and Ricci, E.
\newblock solo-learn: A library of self-supervised methods for visual representation learning.
\newblock \emph{Journal of Machine Learning Research}, 23\penalty0 (56):\penalty0 1--6, 2022.
\newblock URL \url{http://jmlr.org/papers/v23/21-1155.html}.

\bibitem[Deng et~al.(2009)Deng, Dong, Socher, Li, Li, and Fei-Fei]{2009_ImageNet}
Deng, J., Dong, W., Socher, R., Li, L.-J., Li, K., and Fei-Fei, L.
\newblock {ImageNet: A Large-Scale Hierarchical Image Database}.
\newblock In \emph{CVPR}, 2009.

\bibitem[Devlin(2018)]{devlin2018bert}
Devlin, J.
\newblock Bert: Pre-training of deep bidirectional transformers for language understanding.
\newblock \emph{arXiv preprint arXiv:1810.04805}, 2018.

\bibitem[Ermolov et~al.(2021)Ermolov, Siarohin, Sangineto, and Sebe]{2021_ICML_Ermolov}
Ermolov, A., Siarohin, A., Sangineto, E., and Sebe, N.
\newblock Whitening for self-supervised representation learning.
\newblock In \emph{ICML}, 2021.

\bibitem[Geiping et~al.(2022)Geiping, Goldblum, Somepalli, Shwartz-Ziv, Goldstein, and Wilson]{geiping2022much}
Geiping, J., Goldblum, M., Somepalli, G., Shwartz-Ziv, R., Goldstein, T., and Wilson, A.~G.
\newblock How much data are augmentations worth? an investigation into scaling laws, invariance, and implicit regularization.
\newblock \emph{arXiv preprint arXiv:2210.06441}, 2022.

\bibitem[Grill et~al.(2020)Grill, Strub, Altch\'{e}, Tallec, Richemond, Buchatskaya, Doersch, Avila~Pires, Guo, Gheshlaghi~Azar, Piot, kavukcuoglu, Munos, and Valko]{2020_NIPS_Grill}
Grill, J.-B., Strub, F., Altch\'{e}, F., Tallec, C., Richemond, P., Buchatskaya, E., Doersch, C., Avila~Pires, B., Guo, Z., Gheshlaghi~Azar, M., Piot, B., kavukcuoglu, k., Munos, R., and Valko, M.
\newblock Bootstrap your own latent - a new approach to self-supervised learning.
\newblock In \emph{NeuraIPS}, 2020.

\bibitem[Gupta et~al.(2022)Gupta, Ajanthan, Hengel, and Gould]{2022_projector_Gupta}
Gupta, K., Ajanthan, T., Hengel, A. v.~d., and Gould, S.
\newblock Understanding and improving the role of projection head in self-supervised learning.
\newblock \emph{arXiv preprint arXiv:2212.11491}, 2022.

\bibitem[He et~al.(2020)He, Fan, Wu, Xie, and Girshick]{2020_CVPR_He}
He, K., Fan, H., Wu, Y., Xie, S., and Girshick, R.
\newblock Momentum contrast for unsupervised visual representation learning.
\newblock In \emph{CVPR}, 2020.

\bibitem[Huang et~al.(2024)Huang, Ni, Weng, Anwer, Khan, Yang, and Khan]{2024_TPAMI_Huang}
Huang, L., Ni, Y., Weng, X., Anwer, R.~M., Khan, S., Yang, M.-H., and Khan, F.~S.
\newblock Understanding whitening loss in self-supervised learning.
\newblock \emph{IEEE Transactions on Pattern Analysis \& Machine Intelligence}, \penalty0 (01):\penalty0 1--12, 2024.

\bibitem[Hubert \& Arabie(1985)Hubert and Arabie]{1985_JC_Hubert}
Hubert, L. and Arabie, P.
\newblock Comparing partitions.
\newblock \emph{Journal of classification}, 2:\penalty0 193--218, 1985.

\bibitem[Jing et~al.(2022)Jing, Vincent, LeCun, and Tian]{2022_ICLR_Jing}
Jing, L., Vincent, P., LeCun, Y., and Tian, Y.
\newblock Understanding dimensional collapse in contrastive self-supervised learning.
\newblock In \emph{ICLR}, 2022.

\bibitem[Krizhevsky et~al.(2009)Krizhevsky, Hinton, et~al.]{2009_TR_Alex}
Krizhevsky, A., Hinton, G., et~al.
\newblock Learning multiple layers of features from tiny images.
\newblock 2009.

\bibitem[Lin et~al.(2014)Lin, Maire, Belongie, Hays, Perona, Ramanan, Dollar, and Zitnick]{2014_ECCV_COCO}
Lin, T.-Y., Maire, M., Belongie, S., Hays, J., Perona, P., Ramanan, D., Dollar, P., and Zitnick, L.
\newblock Microsoft coco: Common objects in context.
\newblock In \emph{ECCV}, 2014.

\bibitem[Liu et~al.(2022)Liu, Wang, Li, and Wang]{2022_NIPS_Liu}
Liu, X., Wang, Z., Li, Y.-L., and Wang, S.
\newblock Self-supervised learning via maximum entropy coding.
\newblock In Oh, A.~H., Agarwal, A., Belgrave, D., and Cho, K. (eds.), \emph{Advances in Neural Information Processing Systems}, 2022.

\bibitem[Loshchilov \& Hutter(2017)Loshchilov and Hutter]{2017_ICLR_Ilya}
Loshchilov, I. and Hutter, F.
\newblock {SGDR:} stochastic gradient descent with restarts.
\newblock In \emph{ICLR}, 2017.

\bibitem[Ma et~al.(2023)Ma, Hu, and Wang]{2023_arxiv_Ma}
Ma, J., Hu, T., and Wang, W.
\newblock Deciphering the projection head: Representation evaluation self-supervised learning.
\newblock \emph{arXiv preprint arXiv:2301.12189}, 2023.

\bibitem[Morningstar et~al.(2024)Morningstar, Bijamov, Duvarney, Friedman, Kalibhat, Liu, Mansfield, Rojas-Gomez, Singhal, Green, et~al.]{morningstar2024augmentations}
Morningstar, W., Bijamov, A., Duvarney, C., Friedman, L., Kalibhat, N., Liu, L., Mansfield, P., Rojas-Gomez, R., Singhal, K., Green, B., et~al.
\newblock Augmentations vs algorithms: What works in self-supervised learning.
\newblock \emph{arXiv preprint arXiv:2403.05726}, 2024.

\bibitem[Moutakanni et~al.(2024)Moutakanni, Oquab, Szafraniec, Vakalopoulou, and Bojanowski]{moutakanni2024you}
Moutakanni, T., Oquab, M., Szafraniec, M., Vakalopoulou, M., and Bojanowski, P.
\newblock You don't need data-augmentation in self-supervised learning.
\newblock \emph{arXiv preprint arXiv:2406.09294}, 2024.

\bibitem[Nilsback \& Zisserman(2008)Nilsback and Zisserman]{nilsback2008automated}
Nilsback, M.-E. and Zisserman, A.
\newblock Automated flower classification over a large number of classes.
\newblock In \emph{2008 Sixth Indian conference on computer vision, graphics \& image processing}, pp.\  722--729. IEEE, 2008.

\bibitem[Oord et~al.(2018)Oord, Li, and Vinyals]{2018_arxiv_Oord}
Oord, A. v.~d., Li, Y., and Vinyals, O.
\newblock Representation learning with contrastive predictive coding.
\newblock \emph{arXiv preprint arXiv:1807.03748}, 2018.

\bibitem[Oquab et~al.(2023)Oquab, Darcet, Moutakanni, Vo, Szafraniec, Khalidov, Fernandez, Haziza, Massa, El-Nouby, et~al.]{2023_arxiv_Oquab}
Oquab, M., Darcet, T., Moutakanni, T., Vo, H., Szafraniec, M., Khalidov, V., Fernandez, P., Haziza, D., Massa, F., El-Nouby, A., et~al.
\newblock Dinov2: Learning robust visual features without supervision.
\newblock \emph{arXiv preprint arXiv:2304.07193}, 2023.

\bibitem[Papyan et~al.(2020)Papyan, Han, and Donoho]{2020_NAD_Papyan}
Papyan, V., Han, X., and Donoho, D.~L.
\newblock Prevalence of neural collapse during the terminal phase of deep learning training.
\newblock \emph{Proceedings of the National Academy of Sciences}, 117\penalty0 (40):\penalty0 24652--24663, 2020.

\bibitem[Parkhi et~al.(2012)Parkhi, Vedaldi, Zisserman, and Jawahar]{parkhi2012cats}
Parkhi, O.~M., Vedaldi, A., Zisserman, A., and Jawahar, C.
\newblock Cats and dogs.
\newblock In \emph{2012 IEEE conference on computer vision and pattern recognition}, pp.\  3498--3505. IEEE, 2012.

\bibitem[Purushwalkam \& Gupta(2020)Purushwalkam and Gupta]{purushwalkam2020demystifying}
Purushwalkam, S. and Gupta, A.
\newblock Demystifying contrastive self-supervised learning: Invariances, augmentations and dataset biases.
\newblock \emph{Advances in Neural Information Processing Systems}, 33:\penalty0 3407--3418, 2020.

\bibitem[Rousseeuw(1987)]{1987_JCAM_Rousseeuw}
Rousseeuw, P.~J.
\newblock Silhouettes: a graphical aid to the interpretation and validation of cluster analysis.
\newblock \emph{Journal of computational and applied mathematics}, 20:\penalty0 53--65, 1987.

\bibitem[Tian et~al.(2020)Tian, Krishnan, and Isola]{2020_ECCV_Tian}
Tian, Y., Krishnan, D., and Isola, P.
\newblock Contrastive multiview coding.
\newblock In \emph{European conference on computer vision}, 2020.

\bibitem[van~der Maaten \& Hinton(2008)van~der Maaten and Hinton]{Maaten2008VisualizingDU}
van~der Maaten, L. and Hinton, G.~E.
\newblock Visualizing data using t-sne.
\newblock \emph{Journal of Machine Learning Research}, 9:\penalty0 2579--2605, 2008.
\newblock URL \url{https://api.semanticscholar.org/CorpusID:5855042}.

\bibitem[Wagner et~al.(2022)Wagner, Ferreira, Stoll, Schirrmeister, M{\"u}ller, and Hutter]{wagner2022importance}
Wagner, D., Ferreira, F., Stoll, D., Schirrmeister, R.~T., M{\"u}ller, S., and Hutter, F.
\newblock On the importance of hyperparameters and data augmentation for self-supervised learning.
\newblock \emph{arXiv preprint arXiv:2207.07875}, 2022.

\bibitem[Wah et~al.(2011)Wah, Branson, Welinder, Perona, and Belongie]{wah2011caltech}
Wah, C., Branson, S., Welinder, P., Perona, P., and Belongie, S.
\newblock The caltech-ucsd birds-200-2011 dataset.
\newblock 2011.

\bibitem[Wang \& Liu(2021)Wang and Liu]{2021_cvpr_wang_feng}
Wang, F. and Liu, H.
\newblock Understanding the behaviour of contrastive loss.
\newblock In \emph{Proceedings of the IEEE/CVF conference on computer vision and pattern recognition}, pp.\  2495--2504, 2021.

\bibitem[Wang \& Isola(2020)Wang and Isola]{2020_ICML_Wang}
Wang, T. and Isola, P.
\newblock Understanding contrastive representation learning through alignment and uniformity on the hypersphere.
\newblock In \emph{Proceedings of the 37th International Conference on Machine Learning}, 2020.

\bibitem[Weng et~al.(2022)Weng, Huang, Zhao, Anwer, Khan, and Khan]{2022_NIPS_Weng}
Weng, X., Huang, L., Zhao, L., Anwer, R.~M., Khan, S., and Khan, F.
\newblock An investigation into whitening loss for self-supervised learning.
\newblock In \emph{NeurIPS}, 2022.

\bibitem[Weng et~al.(2024)Weng, Ni, Song, Luo, Anwer, Khan, Khan, and Huang]{2024_ICLR_Weng}
Weng, X., Ni, Y., Song, T., Luo, J., Anwer, R.~M., Khan, S., Khan, F.~S., and Huang, L.
\newblock Modulate your spectrum in self-supervised learning.
\newblock In \emph{ICLR}, 2024.

\bibitem[Wu et~al.(2018)Wu, Xiong, Yu, and Lin]{wu2018unsupervised}
Wu, Z., Xiong, Y., Yu, S.~X., and Lin, D.
\newblock Unsupervised feature learning via non-parametric instance discrimination.
\newblock In \emph{Proceedings of the IEEE conference on computer vision and pattern recognition}, pp.\  3733--3742, 2018.

\bibitem[Zbontar et~al.(2021)Zbontar, Jing, Misra, Lecun, and Deny]{2021_NIPS_Zbontar}
Zbontar, J., Jing, L., Misra, I., Lecun, Y., and Deny, S.
\newblock Barlow twins: Self-supervised learning via redundancy reduction.
\newblock In \emph{ICML}, 2021.

\bibitem[Zheng et~al.(2021)Zheng, You, Wang, Qian, Zhang, Wang, and Xu]{2021_NIPS_Zheng}
Zheng, M., You, S., Wang, F., Qian, C., Zhang, C., Wang, X., and Xu, C.
\newblock Ressl: Relational self-supervised learning with weak augmentation.
\newblock \emph{Advances in Neural Information Processing Systems}, 34:\penalty0 2543--2555, 2021.

\end{thebibliography}
\bibliographystyle{icml2025}

%%%%%%%%%%%%%%%%%%%%%%%%%%%%%%%%%%%%%%%%%%%%%%%%%%%%%%%%%%%%%%%%%%%%%%%%%%%%%%%
%%%%%%%%%%%%%%%%%%%%%%%%%%%%%%%%%%%%%%%%%%%%%%%%%%%%%%%%%%%%%%%%%%%%%%%%%%%%%%%
% APPENDIX
%%%%%%%%%%%%%%%%%%%%%%%%%%%%%%%%%%%%%%%%%%%%%%%%%%%%%%%%%%%%%%%%%%%%%%%%%%%%%%%
%%%%%%%%%%%%%%%%%%%%%%%%%%%%%%%%%%%%%%%%%%%%%%%%%%%%%%%%%%%%%%%%%%%%%%%%%%%%%%%

\newpage

\appendix
\onecolumn
\section{Details of Clustering Metrics}
\label{appendix:metrics}
To quantitatively compare the clustering ability of the \Term{encoding}, \Term{embedding}, and the \Term{hidden layer outputs} within the projector, we first define the following metrics in supervised settings (assign samples with the same true label to the same cluster). Let \( \mathcal{X} = \{x_1, x_2, \dots, x_N\} \) be a set of \( N \) data points, and \( \mathcal{Y} = \{y_1, y_2, \dots, y_N\} \) be the corresponding set of true labels.

\begin{definition} (\textbf{Silhouette Coefficient, SC})  
The Silhouette Coefficient~\cite{1987_JCAM_Rousseeuw} is a measure of how similar a sample is to its own cluster compared to its nearest cluster. For a given data point \( x_i \), \( sc(x_i) \) is defined as:

\begin{equation}
    sc(x_i) = \frac{b(x_i) - a(x_i)}{\max(a(x_i), b(x_i))}
\end{equation}

where \( a(x_i) \) is the average distance from point \( x_i \) to all other points \( x_j \) that share the same true label \( y_i = y_j \),
and \( b(x_i) \) is the minimum average distance from point \( x_i \) to all points \( x_j \) that have a different true label \( y_i \neq y_j \). 
\end{definition}
Based on this definition, we know that SC focuses on measuring the local clustering ability of features, with higher values indicating better \textbf{local} clustering ability. Although the SC ranges from $[-1, 1]$, the diversity of features learned through SSL can cause a large value of $\max(a(x_i), b(x_i))$, leading to a smaller effective range for SC. Therefore, in this paper, we note that $\text{SC} > 0$ indicates that the sample has been assigned to the correct cluster. For population statistics, we can compute the mean  $\text{SC}_{\text{mean}}$ and standard deviation $\text{SC}_{\text{std}}$ over all data points in  $\mathcal{X}$.

\begin{definition} (\textbf{Adjusted Rand Index, ARI})
The Adjusted Rand Index~\cite{1985_JC_Hubert} is a measure of the agreement between two partitions of data, adjusted for chance grouping. Given the true labels \( \mathcal{Y} \) and a set of predicted labels \( \mathcal{Y}' = \{ y'_1, y'_2, \dots, y'_N \} \), the ARI is defined as:

\begin{equation}
    \text{ARI} = \frac{\text{RI} - \mathbb{E}[\text{RI}]}{\max(\text{RI}) - \mathbb{E}[\text{RI}]}
\end{equation}

where \( \text{RI} \) is the Rand Index, and \( \mathbb{E}[\text{RI}] \) is its expected value under random labeling.

In practice, the ARI can be computed using a contingency table between \( \mathcal{Y} \) and \( \mathcal{Y}' \). Let \( n_{ij} \) denote the number of data points assigned to the \( i \)-th cluster in \( \mathcal{Y} \) and the \( j \)-th cluster in \( \mathcal{Y}' \). Defining $a_i = \sum_{j} n_{ij}$ and $b_j = \sum_{i} n_{ij}$, then the ARI is calculated as:
\begin{equation*}
    \text{ARI} = \frac{ \sum_{i,j} \binom{n_{ij}}{2} - \left( \sum_i \binom{a_i}{2} \sum_j \binom{b_j}{2} \middle/ \binom{N}{2} \right) }{ \frac{1}{2} \left( \sum_i \binom{a_i}{2} + \sum_j \binom{b_j}{2} \right) - \left( \sum_i \binom{a_i}{2} \sum_j \binom{b_j}{2} \middle/ \binom{N}{2} \right) }
\end{equation*}

where \( \binom{n}{2} = \frac{n(n - 1)}{2} \) is the binomial coefficient. The ARI ranges from $[-0.5, 1]$, where an ARI close to 
$1$ indicates a perfect agreement between the true and predicted labels, and an ARI close to $0$ suggests that the prediction is no better than random assignment.

\end{definition}
Typically, we apply $k$-means ($k$ is set to the true number of classes) clustering on $\mathcal{X}$ to obtain the pseudo labels $\mathcal{Y}'$. Subsequently, ARI is used to measure the agreement between true and pseudo labels, thereby reflecting \textbf{global} clustering ability of the features and the extent to which \textbf{similarity measures} effectively capture the data structure.

\section{Details of Implementation}
\label{appendix:implementation}
In this section, we provide the details and hyperparameters for ReSA pretraining and downstream evaluation.

\subsection{Datasets}
\begin{itemize}
    \item CIFAR-10 and CIFAR-100~\cite{2009_TR_Alex}, two small-scale datasets composed of 32 × 32 images with 10 and 100 classes, respectively. 
    \item ImageNet-100~\cite{2020_ECCV_Tian}, a random 100-class subset of ImageNet~\cite{2009_ImageNet}. 
    \item ImageNet~\cite{2009_ImageNet}, the well-known largescale dataset with about 1.3M training images and 50K test images, spanning over 1000 classes.
    \item COCO2017~\cite{2014_ECCV_COCO}, a large-scale object detection, segmentation, and captioning dataset with 330K images containing 1.5 million object instances.
    \item We also evaluate on fine-grained datasets including CUB-200-2011~\cite{wah2011caltech}, Oxford-IIIT-Pets~\cite{parkhi2012cats}, Food-101~\cite{bossard2014food}, and Oxford-Flowers~\cite{nilsback2008automated}.
\end{itemize}

\vspace{-0.05in}
\subsection{Implementation Details of ReSA Pretraining}
For clarity, we first provide the algorithm of ReSA in PyTorch-style pseudo-code:
\vspace{-0.1in}
\begin{figure}[H]
\includegraphics[width=14cm]{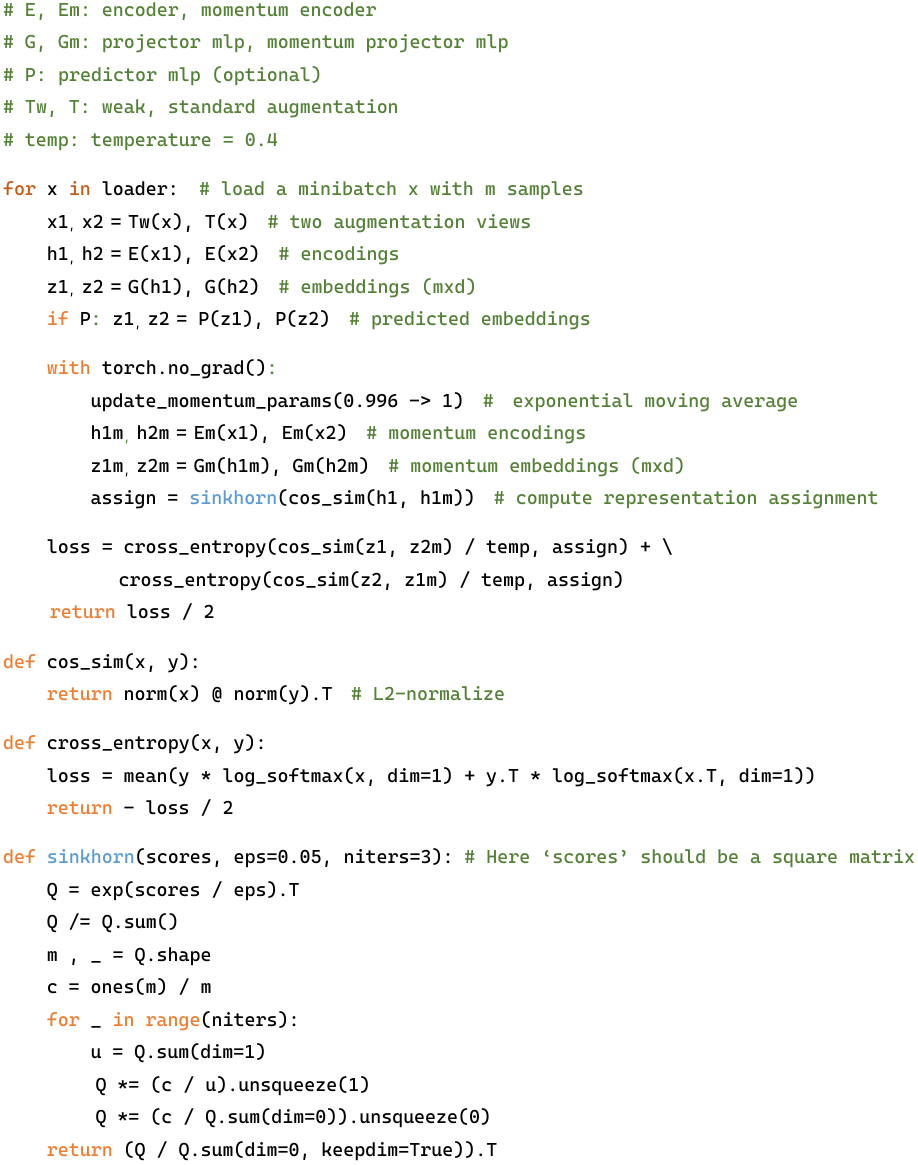}
\end{figure}
\vspace{-0.1in}
\paragraph{Universal settings.} In all experiments conducted in Section~\ref{sec:benchmark_experiments}, we adopt a momentum network, consistent with the practices of most existing self-supervised learning (SSL) methods~\cite{2020_NIPS_Grill,2021_mocov3_Chen,2021_ICCV_Caron,2022_NIPS_Liu,2024_ICLR_Weng}. While the momentum network is not necessary to prevent collapse in ReSA, it has been shown to effectively promote long-term learning in SSL models~\cite{2020_CVPR_He, 2021_CVPR_Chen}. The momentum coefficient, temperature, and Sinkhorn-Knopp parameters in ReSA are configured in accordance with the pseudo-code provided earlier, without requiring further tuning. Furthermore, a standard three-layer MLP is employed as the projector, featuring a hidden layer dimension of 2048 and an output embedding dimension of 512. For training on large-scale datasets, such as ImageNet-1K, we follow the practices of MoCoV3~\cite{2021_mocov3_Chen} and MEC~\cite{2022_NIPS_Liu}, appending a two-layer MLP predictor to the projector in ReSA. The hidden layer and embedding dimensions of the predictor are kept identical to those of the projector. Additionally, we adopt a conventional configuration: when using ConvNets as the encoder, batch normalization (BN) and ReLU activation are incorporated into the hidden layers of both the projector and predictor. However, for ViT-based encoders, we draw inspiration from DINO~\cite{2021_ICCV_Caron}, omitting BN and replacing ReLU with the gaussian error linear units (GELU) activation. This modification ensures that ReSA operates as a \textit{BN-free} system during ViT training, eliminating the need for BN synchronization and offering an improvement in training efficiency.

Building on the aforementioned settings, the only modifications required pertain to optimizer-related parameters, including the learning rate, weight decay, and the number of warmup epochs. These adjustments are made in accordance with the specific encoder architecture and dataset used. Nonetheless, certain settings remain fixed. For instance, we adopt the linear scaling rule, setting the learning rate as
$\textit{lr} = \textit{base lr} \times \textit{batch size} / 256$. After the warmup phase, the learning rate follows a cosine decay schedule~\cite{2017_ICLR_Ilya}.

\paragraph{Details in training ConvNets.} We follow the exact same optimization settings and parameters as INTL~\cite{2024_ICLR_Weng} when pretraining ConvNets. The SGD optimizer is used and detailed parameters are provided in Table~\ref{tab:convnets_param}. The only exception is when training ResNet-50 on ImageNet for 800 epochs, where we reduce the $\textit{base lr}$ to $0.4$ to ensure training stability. Additionally, it is worth noting that we observe a slight performance drop in ReSA when the learning rate decreases to a very small value during the later stages of training. We hypothesize that an excessively small learning rate may amplify the regularization effect of weight decay in SGD, causing the weights to diverge from the optimal solution. To address this, we set the minimum learning rate in the cosine decay schedule to $0.1 * \textit{lr}$.

\begin{table}[H]
 \vspace{-0.1in}
    \centering
    \caption{Optimizer-related parameters in ReSA pretraining.}
    \vspace{0.1in}
    %\footnotesize
    \setlength{\tabcolsep}{6pt}  % Increase column spacing for readability
    \renewcommand{\arraystretch}{1.4}  % Increase row height for better spacing
    \begin{tabular}{ccccccccc}
        \toprule
        Method & dataset & encoder & predictor & optimizer & \textit{batch size}  & \textit{base lr} & weight decay & warmup \\
       \arrayrulecolor{gray}
        \midrule
     \multirow{7}{*}{ReSA} & CIFAR-10 & ResNet-18 & $\usym{2717}$ & SGD & 256  & 0.3 &  $10^{-4}$ &  2 epochs\\
     \cline{2-9}
     & CIFAR-100 & ResNet-18 & $\usym{2717}$ & SGD & 256 & 0.3 & $10^{-4}$ & 2 epochs \\
     \cline{2-9}
     & ImageNet-100 & ResNet-18 & $\usym{2717}$ & SGD & 128  & 0.5 & $2.5 \times 10^{-5}$ & 2 epochs\\
     \cline{2-9}
     & \multirow{3}{*}{ImageNet} & \multirow{2}{*}{ResNet-50} & \multirow{2}{*}{$\checkmark$} & \multirow{2}{*}{SGD} & 256 &  0.5 & $10^{-5}$ & 2 epochs \\
     \cline{6-9}
    & & & &  & 1024 &  0.5 & $10^{-5}$ & 10 epochs\\
    \cline{3-9}
    & & ViT-S/16 & $\checkmark$ & AdamW & 1024 & $5 \times 10^{-4}$ & 0.1 & 40 epochs\\
    \arrayrulecolor{black}
    \bottomrule
    \end{tabular}
     \label{tab:convnets_param}
     \vspace{-0.1in}
\end{table}

\paragraph{Details in training ViTs.} Vision Transformer (ViT) pre-training involves numerous intricate settings, such as initialization methods and optimization parameters, which have a significant impact on training outcomes. Notably, representative SSL methods on ViTs, such as DINO and MoCoV3, adopt totally distinct designs in both architecture and training strategies. DINO~\cite{2021_ICCV_Caron} leverages a range of training tricks, including weight decay scheduling, gradient clipping, and stochastic depth, among others. To stabilize training, it avoids mixed-precision training, which substantially reduces computational efficiency and increases memory requirements. MoCoV3~\cite{2021_mocov3_Chen}, on the other hand, proposes freezing the patch embedding layer to enhance training stability while enabling mixed-precision training. However, it still incorporates batch normalization (BN) in the projector and predictor MLPs, which reduces training efficiency and complicates its application in multi-view scenarios~\cite{morningstar2024augmentations}. Taking these considerations into account, we adopt the ViT design and initialization approach of MoCoV3 for ReSA, but remove the BN layers from the MLPs. This ensures that ReSA functions as a \textit{BN-free} system during ViT training, eliminating the need for BN synchronization and improving overall training efficiency. In this paper, we set the optimizer-related parameters as shown in Table~\ref{tab:convnets_param}. We believe that further exploration of more suitable initialization methods and related parameters for ReSA could enhance its performance, as evidenced by its outstanding results on ConvNets. 

\subsection{Implementation Details of ReSA Evaluating}

\paragraph{Details in evaluating CIFAR-10/100.} 
When evaluating on CIFAR-10/100, we adopt the same linear evaluation protocol as in W-MSE~\cite{2021_ICML_Ermolov} and INTL~\cite{2024_ICLR_Weng}: training a linear classifier for 500 epochs on each labeled dataset using the Adam optimizer, without data augmentation. The learning rate is exponentially decayed from $10^{-2}$ to $10^{-6}$ and the weight decay is $5 \times 10^{-6}$. Under these settings, a single-GPU evaluation takes under one minute—substantially faster than the protocol in solo-learn~\cite{solo_learn_victor}, which can take tens of minutes. We also apply this evaluation protocol to models provided by solo-learn; however, their performance degrades noticeably, so we report the official results in Table~\ref{tab:small_medium_baseline}. In addition, following W-MSE and INTL, we evaluate a simple \textit{5-nn} classifier ($k = 5$) on these datasets for completeness. We track both \textit{linear} and \textit{k-nn} classifier accuracies for ReSA throughout training and observe that at certain checkpoints, ReSA achieves even higher performance than the final values reported in Table~\ref{tab:small_medium_baseline} 
 (e.g., linear accuracies of $93.89\%$ on CIFAR-10 and $72.5\%$ on CIFAR-100). Nevertheless, we report only the final checkpoint’s results for consistency.

\begin{table}[H]
    \centering
     \vspace{-0.1in}
    \caption{Optimal learning rate (\textit{lr}) for training linear classifiers. }
    \vspace{0.1in}
    %\footnotesize
    \setlength{\tabcolsep}{6pt}  % Increase column spacing for readability
    \renewcommand{\arraystretch}{1.4}  % Increase row height for better spacing
    \begin{tabular}{cccc|c}
        \toprule
        \multirow{2}{*}{Method} & \multicolumn{3}{c|}{pretrained settings} & linear eval. \\
        & dataset & encoder & \textit{batch size}  & \textit{lr} \\
       \arrayrulecolor{gray}
        \midrule
     \multirow{4}{*}{ReSA} & ImageNet-100 & ResNet-18 &  128  & 5 \\
     \cline{2-5}
     & \multirow{3}{*}{ImageNet} & \multirow{2}{*}{ResNet-50} &  256 &  10 \\
     \cline{4-5}
    &  &  & 1024 &  40 \\
    \cline{3-5}
    & &  ViT-S/16 & 1024 & 0.03 \\
    \arrayrulecolor{black}
    \bottomrule
    \end{tabular}
     \label{tab:linear_lr}
      \vspace{-0.1in}
\end{table}

\paragraph{Details in evaluating ImageNet-100/1K.} For linear evaluation, we train the linear classifier for 100 epochs with SGD optimizer and using \Term{MultiStepLR} scheduler with $\gamma=0.1$ dropping at the last 40 and 20 epochs. In all our linear classifier training, we fix the batch size at 256 and set the weight decay to 0. However, when using different pretraining datasets, encoders, or batch sizes, the optimal learning rate for evaluation varies accordingly. The specific optimal values for each setting are provided in Table~\ref{tab:linear_lr}. In addition, when training the linear classifier with ViT-S/16, we follow  BERT~\cite{devlin2018bert} and DINO~\cite{2021_ICCV_Caron} by concatenating the [CLS] tokens from the last four layers. 

\begin{table}[ht]
     \vspace{-0.1in}
    \centering
        \caption{Low-shot evaluation. All models are pretrained on ImageNet with ResNet-50, and then fine-tuned with a linear classifer on $1\%$ or $10\%$ subset of ImageNet for 20 epochs. $\dag$ indicates employing the multi-crop trick during pretraining.}
        \vspace{0.1in}
        \setlength{\tabcolsep}{9pt}  % Increase column spacing for readability
        \renewcommand{\arraystretch}{1}  % Increase
        \begin{tabular}{ccccc}
        \toprule
            \multirow{2}{*}{Method} &  \multicolumn{2}{c}{\textbf{top-1}}  & \multicolumn{2}{c}{\textbf{top-5}} \\
            & $1\%$ & $10\%$  &$1\%$ & $10\%$ \\ \midrule
            SimCLR & 48.3&65.6&75.5&87.8 \\
           BYOL &  53.2&68.8&78.4&89.0  \\
            SwAV$^{\dag}$ &53.9&70.2&78.5  &89.9\\
            Barlow Twins & 55.0&69.7&79.2&89.3 \\
            VICReg & 54.8&69.5 &79.4&89.5  \\
           INTL &55.0&69.4&80.8&89.8 \\
            \arrayrulecolor{gray}
            \midrule
            \textbf{ReSA (ours)} & \textbf{56.4} & \textbf{70.4} & \textbf{81.0}   & \textbf{90.1} \\ 
        \arrayrulecolor{black}
        \bottomrule
        \end{tabular}
        \label{tab:semi_super}
        \vspace{-0.05in}
    \end{table}
    
We further evaluate the low-shot learning capability of ReSA in semi-supervised classification. Specifically, we fine-tune the pre-trained ReSA encoder and train a linear classifier for 20 epochs, using $1\%$ and $10\%$ subsets of ImageNet, following the same splits as SimCLR~\cite{2020_arxiv_Chen}. The optimization is conducted using the SGD optimizer with a learning rate of 0.0002 for the encoder and 40 for the classifier, under a batch size of 256, along with a cosine decay schedule. The results, presented in Table~\ref{tab:semi_super}, demonstrate that ReSA also performs effectively in low-shot learning scenarios.

\paragraph{Details in evaluating fine-grained datasets.} In the evaluation experiments on fine-grained datasets presented in Table~\ref{tab:fine_grained}, we apply a weighted \textit{k-nearest neighbors} classifier following~\cite{wu2018unsupervised}. We freeze the pretrained model to compute and store the features of the training data and use these features to select the nearest neighbors for the data in the test set. Based on the top \(k\)-nearest neighbors (\(\mathcal{N}_k\)), predictions are made using a weighted voting mechanism. Specifically, the class \(c\) receives a total weight of $\sum_{i \in \mathcal{N}_k} \alpha_i \mathbf{1}_{c_i = c}$,
where \(\alpha_i\) represents the contribution weight. We compute \(\alpha_i = \exp\left(\frac{T_i \cdot x}{\tau}\right)\), with \(\tau\) set to 0.07 as described in ~\cite{wu2018unsupervised} and used by DINO, without tuning this value. 

%%%%%%%%%%%%%%%%%%%%%%%%%%%%%%%%%%%%%%%%%%%%%%%%%%%%%%%%%%%%%%%%%%%%%%%%%%%%%%%
%%%%%%%%%%%%%%%%%%%%%%%%%%%%%%%%%%%%%%%%%%%%%%%%%%%%%%%%%%%%%%%%%%%%%%%%%%%%%%%
 \begin{figure*}[t]
    %\vskip 0.2in
    \begin{center}
    \centerline{\includegraphics[width=16cm]{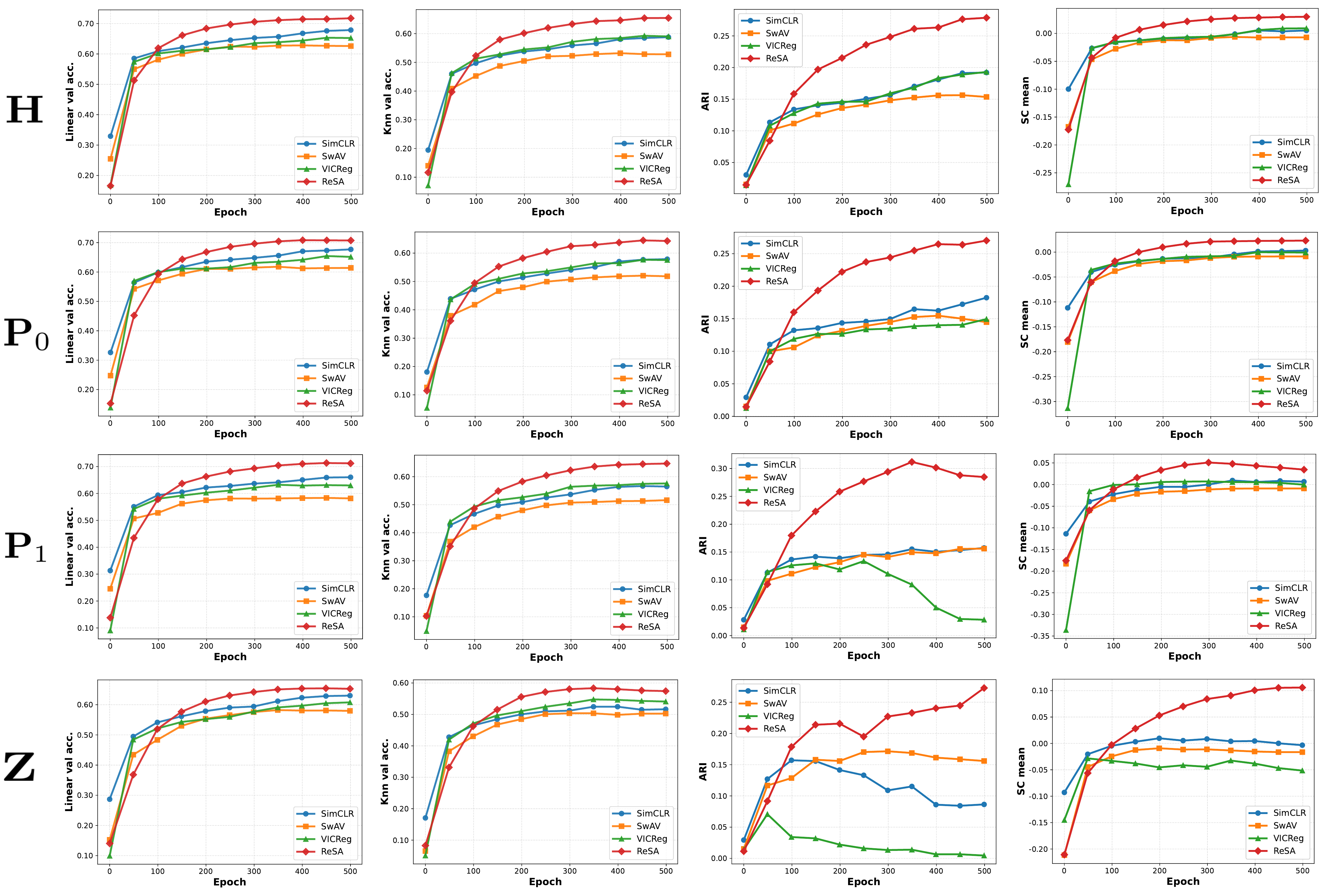}}
    \vspace{-0.1in}
    \caption{Comparison of evaluation accuracies and clustering metrics among various SSL methods during the training process. The experiments are conducted using SimCLR, SwAV, VICReg, and ReSA. The settings and notations are consistent with ones in Figure~\ref{fig:rep_cifar100}.}
    \label{fig:cifar100_compare}
    \end{center}
    \vspace{-0.3in}
    \end{figure*}

\vspace{-0.1in}
\section{Additional analyses on ReSA} 
\label{sec:ablation}

\subsection{Ablation Studies} 
As presented in Table~\ref{tab:ablation}, we conduct ablation studies to investigate the impact of network architecture design and temperature hyperparameter selection on ReSA performance. Our analysis reveals that employing a momentum network yields an accuracy improvement of approximately $2\%$, albeit at the cost of increased computational overhead. In contrast, the integration of an additional predictor demonstrates a comparable accuracy gain of around $1\%$ while maintaining near-identical computational efficiency, exhibiting negligible impact on runtime performance.

Meanwhile, in contrast to contrastive learning methods and other approaches such as SwAV and DINO, which typically require a small temperature value (e.g. $\tau =0.1$), ReSA achieves favorable performance with a higher temperature value ($\tau =0.4$). This indicates that the optimization process of ReSA can effectively incorporate a broader range of samples with better tolerance, rather than focusing exclusively on hard samples~\cite{2021_cvpr_wang_feng}, as is the case with other methods.

\begin{figure*}[ht]
    %\vskip 0.2in
    \begin{center}
    \centerline{\includegraphics[width=17cm]{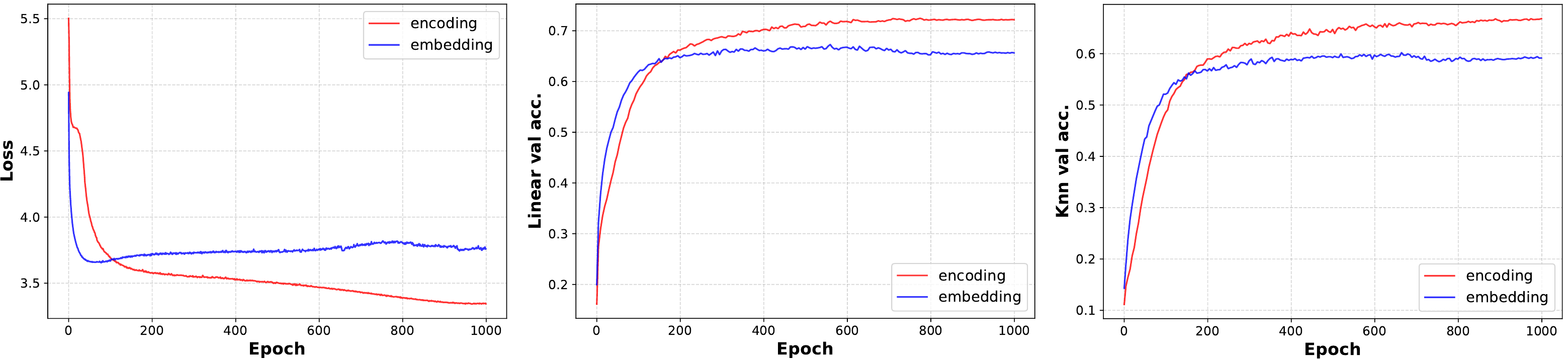}}
    \vspace{-0.1in}
    \caption{Ablation on extraction of clustering information from \Term{encoding} vs. \Term{embedding} to obtain the self-assignment $\MA{\MH{}}$. Both are pretrained for 1000 epochs on CIFAR-100 using ResNet-18 under totally the same experimental settings provided in Appendix~\ref{appendix:implementation}.}
    \label{fig:emb}
    \end{center}
    \end{figure*}

Additionally, we conduct experiments to test the extraction of clustering information from \Term{embedding} to obtain the self assignment $\MA{\MH{}}$. The results shown in Figure~\ref{fig:emb} indicate that, under this condition, the loss struggles to converge, and the model's accuracy significantly declines compared to ReSA. This finding is consistent with the analyses in Section~\ref{sec:properties}, where we note that the clustering properties of \Term{embedding} are less stable and inferior to those of \Term{encoding}, making it challenging for the model to effectively learn high-quality clustering information.
\begin{table}[H]
    \centering
    \caption{Ablation studies on the network architecture and temperature hyperparameter of ReSA on ImageNet using ResNet-50 as the encoder. When evaluating the impact of different network architectures on ReSA, we set the batch size to 256 and perform pretraining using a single GPU.}
    \vspace{0.1in}
    %\footnotesize
    \setlength{\tabcolsep}{7pt}  % Increase column spacing for readability
    \renewcommand{\arraystretch}{1.4}  % Increase row height for better spacing
    \begin{tabular}{ccccccc}
        \toprule
        Method & momentum & predictor & \textit{linear acc.} & memory (GB) & time (h/epoch) \\
       \arrayrulecolor{gray}
        \midrule
     \multirow{4}{*}{ReSA} & $\checkmark$ & $\checkmark$ & \textbf{71.9} &  25.2 &   0.78 &\\
     \cline{2-6}
     & $\checkmark$ & $\usym{2717}$ & 70.8 & 25.2 &  0.78 & \\
     \cline{2-6}
     & $\usym{2717}$ & $\checkmark$ & 69.7 & 24.3 &  0.58 &\\
     \cline{2-6}
     & $\usym{2717}$ & $\usym{2717}$ & 68.7 &  24.3 &  0.58 &\\
    \arrayrulecolor{black}
    \bottomrule
    \end{tabular}
    \\[0.5cm]
    \begin{tabular}{cccccc}
       \toprule
        %\noalign{\smallskip}
       \multirow{2}{*}{Method} & \multirow{2}{*}{batch size} & \multicolumn{4}{c}{temperature $\tau$} \\
       & & 0.2 & 0.3 & 0.4 & 0.5\\
        \arrayrulecolor{gray}
        \midrule
        \multirow{2}{*}{ReSA} & 256  &71.2 & 71.4& \textbf{71.9}& 71.7 \\
        \cline{2-6}
        & 1024  &70.9 & 71.1& \textbf{71.3}& 71.2 \\
       \bottomrule
    \end{tabular}
    \label{tab:ablation}
\end{table}

\begin{figure}[ht]
    \vspace{-0.1in}
    %\vskip 0.2in
    \begin{center}
    \centerline{\includegraphics[width=17cm]{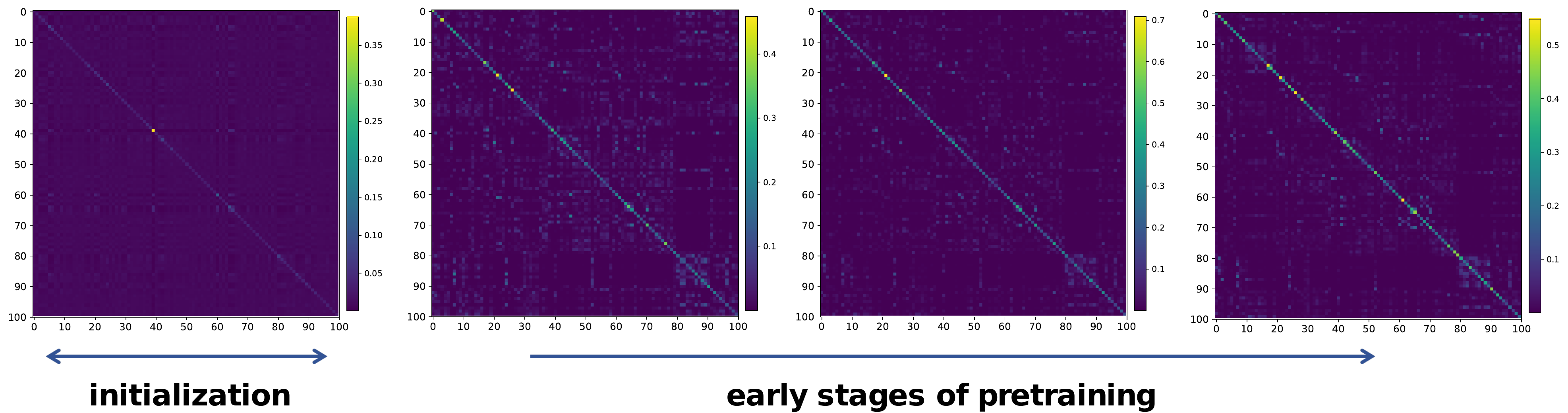}}
    \vspace{-0.1in}
    \caption{Visualization of the self-assignment matrix $\MA{\MH{}}$ during the early stages of training.}
    \label{fig:assign}
    \end{center}
    \vspace{-0.2in}
\end{figure}

\subsection{How ReSA Avoids Feature Collapse and Early Clustering Error?}

Although ReSA has demonstrated good performance in various experiments, it remains unclear how the model avoids feature collapse and clustering error in the early stages of training when it has not yet learned any clustering information.

To answer these questions, we first consider the loss formula of ReSA: $\ell_{\text{ReSA}}(\z{i}) = - \sum_{j=1}^m {\mathbf{A}_\mathbf{H}}^{(i,j)} \log \left( \frac{\exp(s_{i,j} / \tau)}{\sum_{k=1}^m \exp(s_{i,k} / \tau)} \right)$. Here, ${\mathbf{A} _\mathbf{H}}^{(i,j)}$ denotes the $ (i,j) $-th element of $ \mathbf{A} _\mathbf{H} $, where $ \mathbf{A} _\mathbf{H} = \text{Sinkhorn}(\mathbf{H}^\top \mathbf{H}) $. Given that the Sinkhorn-Knopp algorithm exhibits strict monotonicity, we have the property that if $ {\h {i}}^\top {\h {j}} > {\h {i}}^\top {\h {k}}$, then $ {\mathbf{A} _\mathbf{H}}^{(i,j)} > {\mathbf{A} _\mathbf{H}}^{(i,k)}$. Since the vectors in $ \mathbf{H} $ are $L _2 $-normalized, the diagonal elements of $\mathbf{H}^\top \mathbf{H}$ are all maximized to 1. This implies that for any $ i $ and $j $, we have $ {\mathbf{A} _\mathbf{H}}^{(i,i)} \geq {\mathbf{A} _\mathbf{H}}^{(i,j)} $. Furthermore, due to the sharp distribution employed in Sinkhorn, the diagonal elements are significantly larger than the off-diagonal elements. This ensures that the optimization focus of ReSA remains on the alignment of augmented views from the same image, substantially reducing the impact of early assignment errors (off-diagonal elements) on the training process. Throughout training, the continual alignment enables the model to learn meaningful representations, which in turn facilitates correct cluster assignments in the later stages, further promoting learning. In contrast, other clustering-based methods such as SwAV and DINO require an initialized prototype for cluster assignment, making it more challenging to avoid early clustering errors. We speculate that this is one of the key reasons why ReSA outperforms these methods in terms of overall effectiveness.

We then visualize the self-assignment matrix $\MA{\MH{}}$ during the early stages of training. As shown in Figure~\ref{fig:assign}, at model initialization, the assignment already exhibits a dominance of diagonal elements, reflecting an optimization trend that pulls augmented views from the same image closer together. As training progresses, we observe that the dominance of diagonal elements gradually strengthens, effectively preventing feature collapse in the model.

\subsection{Pretraining on Long-tailed Dataset}

\begin{figure}[ht]
    \vspace{-0.1in}
    %\vskip 0.2in
    \begin{center}
    \centerline{\includegraphics[width=17cm]{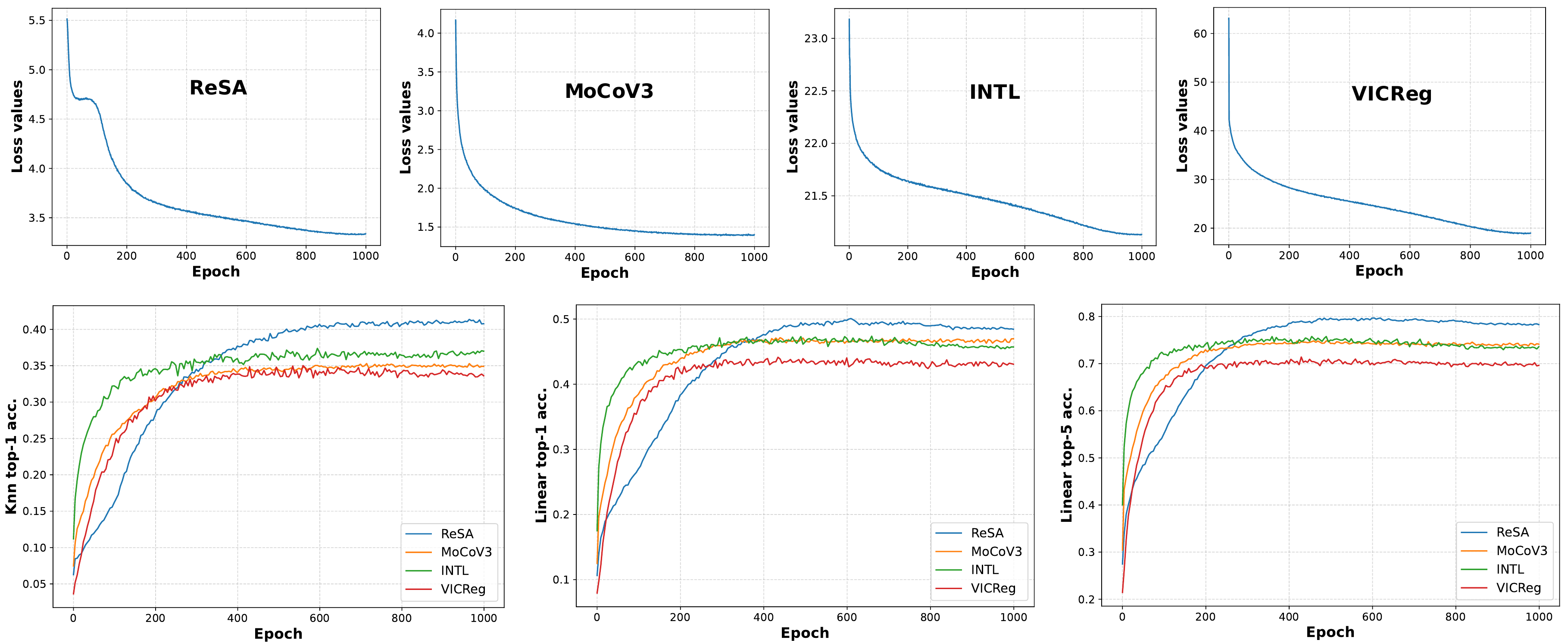}}
    \vspace{-0.1in}
    \caption{Pretraining on the long-tailed dataset. Here We report the training losses of four self-supervised learning methods on the CIFAR100-LT dataset, along with their evalutaion performance on the full CIFAR-100 test set as measured by a linear probe and a k-NN classifier.}
    \label{fig:long_tail}
    \end{center}
    \vspace{-0.2in}
\end{figure}

To further evaluate the training performance of ReSA on imbalanced datasets, we conduct experiments using four self-supervised learning methods—ReSA, MoCoV3, INTL, and VICReg—pretrained and evaluated on the long-tailed CIFAR100-LT dataset. We follow the setup on \texttt{https://huggingface.co/datasets/tomas-gajarsky/cifar100-lt}, setting an imbalance factor of 1/20, resulting in the CIFAR100-LT dataset containing 15,907 images. We train these four SSL models for 1000 epochs with ResNet-18 as the encoder and evaluate it on the full CIFAR100 test set. As observed in Figure~\ref{fig:long_tail}, the loss values of these four methods converge well, but the final evaluation accuracy is significantly lower compared to training on the full CIFAR100 dataset in Table~\ref{tab:small_medium_baseline}. Interestingly, we notice that ReSA's loss decreases more slowly in the early stages, and its accuracy improves more gradually than other methods. We hypothesize that this may be due to noisy initial clusters in the early stages of training, causing clustering errors. However, we find that as training progresses, ReSA's accuracy continues to rise in the mid-phase, surpassing all other methods. This suggests that ReSA is able to gradually resolve these issues and learn the correct clustering patterns as training advances, rather than amplifying errors. Overall, this experiment demonstrates that ReSA can also learn effective representations on long-tailed datasets.

\subsection{The impact of weak augmentation on other methods}
\label{sec:weak_aug}

Since the design of weak augmentation (identical to that in ReSSL~\cite{2021_NIPS_Zheng}) provides a certain degree of performance improvement for ReSA, we also examine its impact on other self-supervised learning methods. Specifically, we select SwAV, VICReg, and MoCoV3, strictly following the experimental configurations in solo-learn~\cite{solo_learn_victor}, with the only modification being the replacement of the image augmentation settings. As shown in Table~\ref{tab:weak_aug}, weak augmentation do not enhance the performance of these methods. This is likely because the standard augmentation settings have been extensively tuned through numerous experiments, ensuring optimal training conditions for these approaches.

\begin{table}[H]
    \centering
    \caption{Investigate the impact of weak augmentation on other methods.}
    \vspace{0.1in}
    %\footnotesize
    \setlength{\tabcolsep}{8pt}  % Increase column spacing for readability
    \renewcommand{\arraystretch}{1.2}  % Increase row height for better spacing
    \begin{tabular}{cccccc}
       \toprule
        %\noalign{\smallskip}
       \multirow{2}{*}{Method} & \multirow{2}{*}{weak aug.} & \multicolumn{2}{c}{CIFAR-10} & \multicolumn{2}{c}{CIFAR-100}\\
       & & \textit{linear} & \textit{k-nn} & \textit{linear} & \textit{k-nn} \\
        \arrayrulecolor{gray}
        \midrule
        \multirow{2}{*}{SwAV} & $\usym{2717}$  & \textbf{89.17} & \textbf{84.18} & \textbf{64.88} & \textbf{53.32} \\
        & $\checkmark$  & 88.79 & 84.01& 64.21 & 53.07 \\
        \arrayrulecolor{gray}
            \midrule
        \multirow{2}{*}{VICReg} & $\usym{2717}$  & \textbf{92.07} & \textbf{87.38} & \textbf{68.54}& \textbf{56.32} \\
        & $\checkmark$  &91.35 & 86.75& 67.27 & 55.79 \\
        \arrayrulecolor{gray}
            \midrule
        \multirow{2}{*}{MoCoV3} & $\usym{2717}$  &\textbf{93.10} & 89.06 & \textbf{68.83} & \textbf{58.09} \\
        & $\checkmark$  & 93.05 & \textbf{89.09} & 68.72 & 58.02 \\
       \bottomrule
    \end{tabular}
    \label{tab:weak_aug}
\end{table}

\subsection{Discussion and Future Work}

The relationship between image augmentation and SSL via joint embedding architectures has grown increasingly intricate. Over the past few years, many studies~\cite{2020_arxiv_Chen,2020_NIPS_Grill,wagner2022importance,morningstar2024augmentations} have emphasized the critical role of image augmentation in JEA, demonstrating that making subtle modifications to image augmentations, such as merely adjusting the parameters of \textit{ResizedCrop} and \textit{ColorJitter}, can significantly impact the performance of SSL models. However, recent works~\cite{2023_cvpr_Assran,moutakanni2024you} have begun to challenge this paradigm by exploring new self-supervised learning frameworks that eliminate the reliance on hand-crafted data augmentations. These efforts argue that specific augmentations may introduce strong biases that could be detrimental to certain downstream tasks~\cite{assran2022hidden} and that the most effective augmentations are often task-specific, depending on the domain, rather than adhering to universally hand-crafted settings~\cite{bendidi2023no,asano2019critical,geiping2022much,purushwalkam2020demystifying}.

Interestingly, \citet{moutakanni2024you} successfully demonstrate that hand-crafted or domain-specific data augmentations are not essential for training state-of-the-art joint embedding architectures when scaling self-supervised learning. Their findings reveal that, with sufficiently large datasets, simple crop of images alone can achieve remarkable results. Notably, this observation aligns perfectly with the characteristics of ReSA. As we show in Section~\ref{sec:aug}, removing any single transformation enhances the clustering properties of the representations learned during training, enabling ReSA to better capture the inherent clustering information within the data. When the dataset size is sufficiently large, eliminating all hand-crafted data augmentations perfectly unleashes the innate potential of ReSA. We look forward to future research validating this hypothesis and applying ReSA to large-scale pretraining scenarios.

\end{document}